\documentclass[lettersize,journal]{IEEEtran}
\usepackage{amsmath,amsfonts}
\usepackage{algorithmic}
\usepackage{algorithm}
\usepackage{array}
\usepackage{amssymb}
\DeclareMathOperator*{\argmin}{argmin}
\usepackage[caption=false,font=normalsize,labelfont=sf,textfont=sf]{subfig}
\usepackage{textcomp}
\usepackage{stfloats}
\usepackage{url}
\usepackage{verbatim}
\usepackage[table]{xcolor}
\usepackage{graphicx}
\usepackage{cite}
\usepackage{makecell}
\usepackage{multirow}
\usepackage{threeparttable}
\usepackage[normalem]{ulem}
\usepackage{overpic}
\usepackage{hyperref}
\usepackage{footnote}
\makesavenoteenv{table}
\useunder{\uline}{\ul}{}
\usepackage{bm}
\usepackage{xspace}
\usepackage{setspace}
\usepackage{dsfont}
\usepackage[cal=cm, scr=rsfs]{mathalpha}
\usepackage{amsmath}
\usepackage{booktabs}
\usepackage[normalem]{ulem}
\useunder{\uline}{\ul}{}
\usepackage{xcolor}         
\usepackage{pifont}

\definecolor{lightcoral}{rgb}{0.94, 0.5, 0.5}
\definecolor{lightgreen}{rgb}{0.56, 0.93, 0.56}
\definecolor{harvestgold}{rgb}{0.85, 0.57, 0.0}
\definecolor{brightlavender}{rgb}{0.75, 0.58, 0.89}
\definecolor{capri}{rgb}{0.0, 0.75, 1.0}
\definecolor{carminepink}{rgb}{0.92, 0.3, 0.26}
\definecolor{celadon}{rgb}{0.67, 0.88, 0.69}
\definecolor{darkpastelgreen}{rgb}{0.01, 0.75, 0.24}

\definecolor{deepred}{rgb}{0.698,0.133,0.133}
\definecolor{blue}{rgb}{0,0,1}
\usepackage{tabularx}
\usepackage{ragged2e}

\makeatletter
\DeclareRobustCommand\onedot{\futurelet\@let@token\@onedot}
\def\@onedot{\ifx\@let@token.\else.\null\fi\xspace}

\def\eg{\emph{e.g}\onedot} 
\def\ie{\emph{i.e}\onedot} 
 
\def\etc{\emph{etc}\onedot} 
 
\def\etal{\emph{et al}\onedot}
\makeatother

\begin{document}
\title{Open-world Machine Learning: A Review and \\New Outlooks}

\author{Fei Zhu$^*$, Shijie Ma$^*$, Zhen Cheng,
	Xu-Yao Zhang~\IEEEmembership{Senior Member,~IEEE},
	Zhaoxiang Zhang~\IEEEmembership{Senior Member},\\
        Dacheng Tao~\IEEEmembership{Fellow,~IEEE} and
	Cheng-Lin Liu~\IEEEmembership{Fellow,~IEEE}
	\IEEEcompsocitemizethanks{
		\IEEEcompsocthanksitem Fei Zhu is with the Centre for Artificial Intelligence and Robotics, Hong Kong Institute of Science and Innovation, Chinese Academy of Sciences, Hong Kong 999077, China.
		\IEEEcompsocthanksitem Shijie Ma, Zhen Cheng, Xu-Yao Zhang, Zhaoxiang Zhang and Cheng-Lin Liu are with the State Key Laboratory of Multimodal Artificial Intelligence Systems, Institute of Automation of Chinese Academy of Sciences, 95 Zhongguancun East Road, Beijing 100190, P.R. China, and also with the School of Artificial Intelligence, University of Chinese Academy of Sciences, Beijing 100049, P.R. China.
        \IEEEcompsocthanksitem Dacheng Tao is with College of Computing and Data Science, Nanyang Technological University, Singapore, Singapore.
        \IEEEcompsocthanksitem Email: fei.zhu@cair-cas.org.hk, \{mashijie2021, chengzhen2019, zhaoxiang.zhang\}@ia.ac.cn, \{xyz, liucl\}@nlpr.ia.ac.cn, dacheng.tao@ntu.edu.sg}
}

\markboth{}%
{Shell \MakeLowercase{\textit{et al.}}: A Sample Article Using IEEEtran.cls for IEEE Journals}

\maketitle

\begin{abstract}
Machine learning has achieved remarkable success in many applications. However, existing studies are largely based on the closed-world assumption, which assumes that the environment is stationary, and the model is fixed once deployed. In many real-world applications, this fundamental and rather naive assumption may not hold because an open environment is complex, dynamic, and full of unknowns. In such cases, rejecting unknowns, discovering novelties, and then continually learning them, could enable models to be safe and evolve continually as biological systems do.
This article presents a holistic view of open-world machine learning by investigating unknown rejection, novelty discovery, and continual learning in a unified paradigm. The challenges, principles, and limitations of current methodologies are discussed in detail. Furthermore, widely used benchmarks, metrics, and performances are summarized. Finally, we discuss several potential directions for further progress in the field. 
By providing a comprehensive introduction to the emerging open-world machine learning paradigm, this article aims to help researchers build more powerful AI systems in their respective fields, and to promote the development of artificial general intelligence.
\end{abstract}

\renewcommand{\thefootnote}{\fnsymbol{footnote}}
\footnotetext[1] {These authors contributed equally to this work.}

\begin{IEEEkeywords}
Machine learning, deep learning, open-world, unknown rejection, novel class discovery, continual learning.
\end{IEEEkeywords}

\section{Introduction}\label{sec:introduction}
Artificial intelligence, coupled with machine learning techniques \cite{bishop2006pattern, li2020accelerated}, is broadly used in many fields, such as medical treatment \cite{acosta2022multimodal}, industry \cite{long2022machine}, transportation and scientific discovery \cite{wang2023scientific}. Typically, supervised machine learning involves isolated classification or regression task, which learns a function (model) $f: \mathcal{X} \rightarrow \mathcal{Y}$ from a training dataset $\mathcal{D} = \{(\bm{x}_{i}, y_{i})\}^{N}_{i=1}$  containing pairs of feature vector and ground-truth label \cite{hastie2009elements}. Then, the model $f$ can be deployed to predict future encountered inputs. 
However, the current success of machine learning is largely based on the closed-world assumption \cite{chen2018lifelong, bendale2015towards, zhang2020towards}, where the important factors of learning are limited to what has been observed during training. In the classification task, all the classes $y$ that the model will encounter during deployment must have been seen in training, \ie, $y \in \mathcal{Y}$. This assumption is often reasonable in restricted scenarios where possible classes are well-defined and unlikely to change over time. For example, in a handwritten digit recognition task, the closed-world assumption holds because the set of digits (0-9) is fixed and known in advance. Besides, this assumption also makes the data collection process easier and straightforward.
\begin{figure}[!t]
\centering
\includegraphics[width=0.5\textwidth]{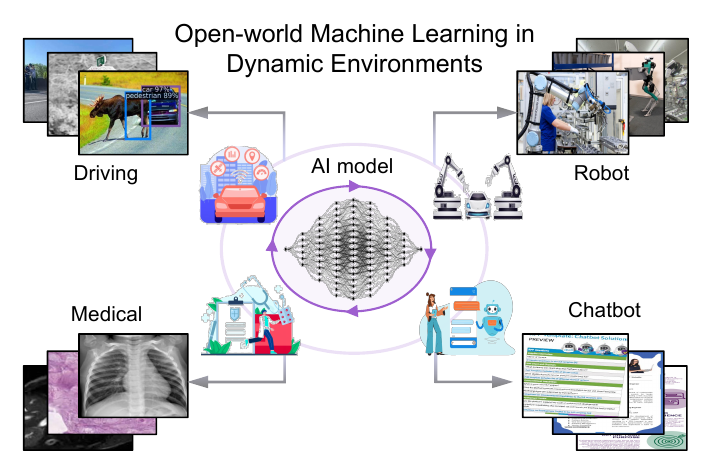}
\vskip -0.1in
\caption{Illustration of open-world machine learning scenarios. In intelligent driving, robot manipulation, medical diagnosis and Chatbot scenarios, the environments are open, dynamic and complex. Open-world machine learning enables the model to learn and evolve from streaming data safely. }
\label{Figure:intro}
\vskip -0.05in
\end{figure}

\begin{figure*}[t]
	\begin{center}
		\centerline{\includegraphics[width=0.95\textwidth]{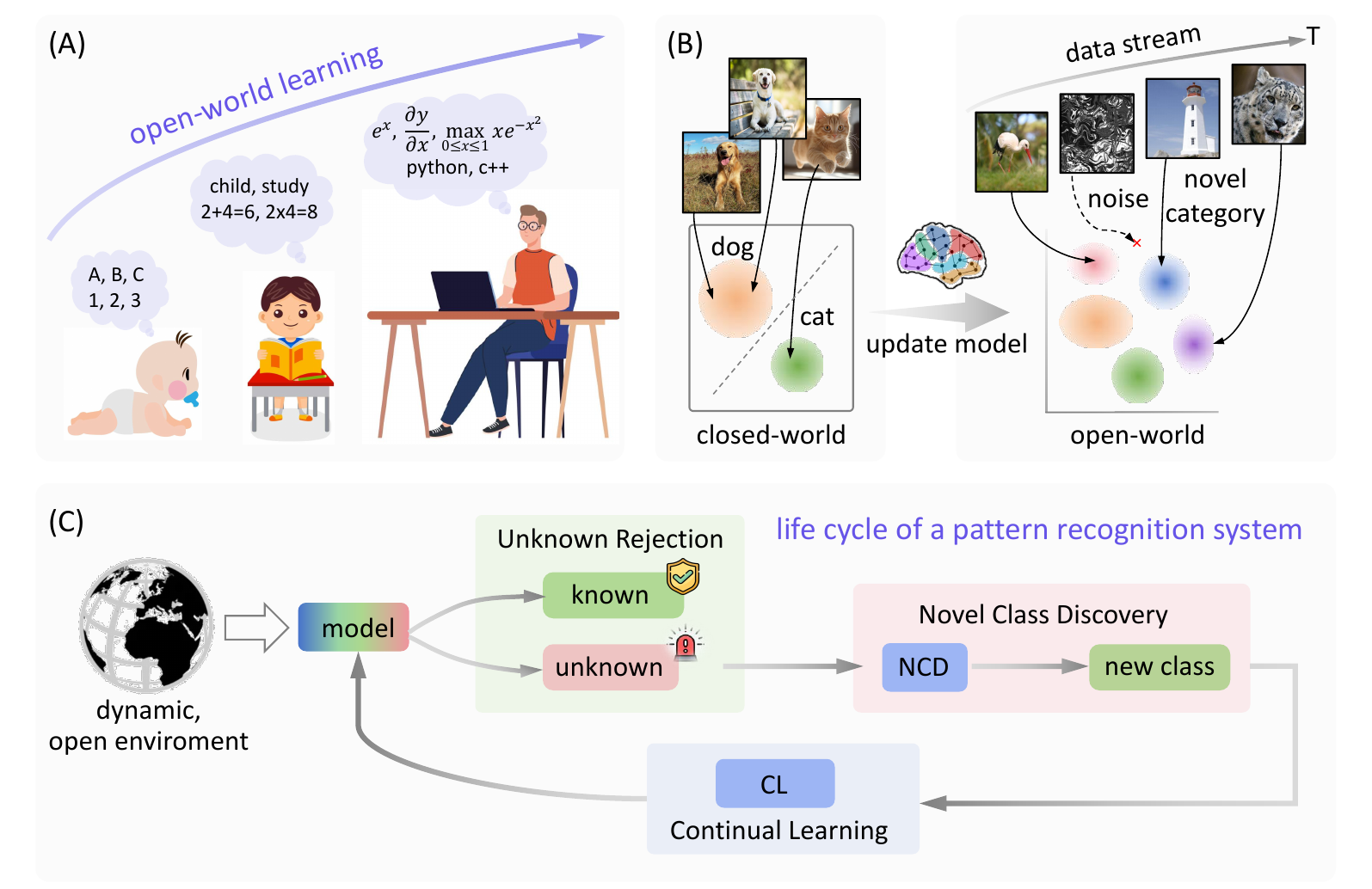}}
		\vskip -0.2in
		\caption{Illustrations of the life cycle of a learning system in the open-world applications. (A) Humans continually learn new knowledge throughout their lives and maintain/use previous knowledge, becoming increasingly smarter and more skillful over time. (B) Open-world machine learning aims to build a human-like system that can transfer and consolidate knowledge continually during deployment. (C) An open-world learning paradigm mainly includes three parts, \ie,  unknown rejection, novel class discovery and continual learning.}
		\label{figure-2}
	\end{center}
	\vskip -0.1in
\end{figure*}

\begin{table*}[!t]
\renewcommand\arraystretch{1.5}
\caption{A comprehensive taxonomy of open-world machine learning, including all three tasks of unknown rejection, novel class discovery and continual learning. Each task is divided into several subcategories with representative methods.}
\vspace{-8pt}
\label{tab:summary}
\begin{center}
    \resizebox{\linewidth}{!}{
        \begin{tabular}{c|c|c|c}
                \hline
                \multicolumn{3}{c|}{Categories} & Representative Methods \\
                \hline
                \hline
                \multirow{8}{*}{\makecell{Unknown Rejection}} & \multirow{4}{*}{\makecell{Out-of-distribution \\Detection}} &
                Post-hoc Inference Methods & \makecell{MSP~\cite{hendrycks2016baseline}, ODIN~\cite{liang2018enhancing}, Energy Score~\cite{liu2020energy},  Mahalanobis~\cite{lee2018simple},\\ViM~\cite{wang2022vim}, WDiscOOD~\cite{chen2023wdiscood}, KNN~\cite{sun2022out}, GEN~\cite{liu2023gen}, Cosine~\cite{noh2023simple}, \\ASH~\cite{djurisic2022extremely}, SCALE~\cite{xu2023scaling}, MOS~\cite{huang2021mos}, Maxlogits~\cite{hendrycks2022scaling} 
                } \\
                \cline{3-4}
                & & Training-stage Methods & \makecell{SSL~\cite{hendrycks2019using}, LogitNorm~\cite{wei2022mitigating}, FMFP~\cite{zhu2023revisiting}, classAug~\cite{zhu2022learning}, AoP \cite{cheng2025average}, \\OVA-PL\cite{cheng2023unified}, RCL \cite{zhu2024rcl}, GenData~\cite{cheng2024breaking}, PixMix \cite{hendrycks2022pixmix}
                } \\
                \cline{3-4}
                & & Outlier-aided Methods & \makecell{OE~\cite{hendrycks2018deep}, OpenMix~\cite{zhu2023openmix}, VOS~\cite{du2022vos}, POE~\cite{ma2025towards}, LoCoOp \cite{miyai2023locoop}, \\NegPrompt \cite{li2024learning}, NegLabel \cite{jiang2024negative}, LocalPrompt~\cite{zeng2025local}, NPOS \cite{tao2023non}} \\
                \cline{2-4}
                & \multirow{3}{*}{\makecell{Open-Set Recognition}} & Discriminative Models & \makecell{OpenMax~\cite{bendale2016towards}, PROSER~\cite{zhou2021learning}, C2AE~\cite{oza2019c2ae}, RAD~\cite{fu2025reason}, ODL~\cite{liu2022orientational}
                } \\
                \cline{3-4}
                & & Generative Models & \makecell{G-OpenMax~\cite{ge2017generative},  OpenGAN~\cite{kong2021opengan}} \\
                \cline{3-4}
                & & Hybrid Models & \makecell{OpenHybrid~\cite{zhang2020hybrid}, CPN~\cite{yang2020convolutional}, CAC \cite{miller2021class}, \\ OVA-PL\cite{cheng2023unified}, ARPL~\cite{chen2021adversarial}, CSSR~\cite{huang2022class}} \\
                \hline
                \multirow{11}{*}{\makecell{Novel Class Discovery}} & \multirow{3}{*}{\makecell{Novel Category \\ Discovery (NCD)}}
                & Isolated-training Methods & \makecell{KCL~\cite{hsu2018learning}, MCL~\cite{hsu2018multiclass}, DTC~\cite{han2019learning}, ResTune~\cite{liu2022residual}} \\
                \cline{3-4}
                & &  Joint-training Methods & \makecell{RandStats~\cite{Han2020Automatically,han2021autonovel}, NCL~\cite{Zhong_2021_CVPR}, OpenMix~\cite{zhong2021openmix}, UNO~\cite{Fini_2021_ICCV}, \\DualRS~\cite{zhao2021novel}, ComEx~\cite{yang2022divide}, IIC~\cite{Li_2023_CVPR}, rKD~\cite{guclass}, TIDA~\cite{wang2023discover}, SCKD~\cite{wang2024self}} \\
                \cline{3-4}
                & &  Continual NCD & \makecell{FRoST~\cite{roy2022class}, NCDwF~\cite{joseph2022novel}, KTRFR~\cite{liu2024large}, ADM~\cite{chen2024adaptive}} \\
                \cline{2-4}
                & \multirow{4}{*}{\makecell{Generalized Category \\ Discovery (GCD)}}
                & Non-parametric Classification & \makecell{GCD~\cite{vaze2022gcd}, XCon~\cite{fei2022xcon}, DCCL~\cite{pu2023dynamic}, GPC~\cite{Zhao_2023_ICCV}, PromptCAL~\cite{zhang2023promptcal}, \\InfoSieve~\cite{rastegar2023learn}, CiPR~\cite{hao2024cipr}, CMS~\cite{choi2024contrastive}} \\
                \cline{3-4}
                & & Parametric Classification & \makecell{ORCA~\cite{cao2022openworld}, PIM~\cite{chiaroni2023parametric}, SimGCD~\cite{Wen_2023_ICCV}, $\mu$GCD~\cite{vaze2023no}, \\SPTNet~\cite{wang2024sptnet}, LegoGCD~\cite{cao2024solving}, ProtoGCD~\cite{10948388}} \\
                \cline{3-4}
                & & Continual GCD & \makecell{GM~\cite{zhang2022grow}, IGCD~\cite{Zhao_2023_ICCV_Incremental}, PA-GCD~\cite{Kim_2023_ICCV}, MetaGCD~\cite{wu2023metagcd}, \\PromptCCD \cite{cendra2024promptccd}, Happy~\cite{NEURIPS2024_5ae0f7cf}} \\
                \cline{2-4}
                & \multirow{5}{*}{\makecell{Extended Settings}}
                & Other Learning Paradigms & \makecell{Active Learning (ActiveGCD~\cite{ma2024active}), Federated Learning (FedGCD~\cite{pu2024federated})\\ Domain Adaptation (HiLo~\cite{wang2025hilo})} \\
                \cline{3-4}
                & & On-the-fly Discovery & SMILE~\cite{du2023fly}, PHE~\cite{zheng2024prototypical} \\
                \cline{3-4}
                & & Fine-grained Discovery & RAPL~\cite{liu2024novel}, SelEx~\cite{rastegar2024selex}, PHE~\cite{zheng2024prototypical}, FineR~\cite{liu2024democratizing} \\
                \cline{3-4}
                & & Long-tailed Discovery & \makecell{ImbaGCD~\cite{li2023imbagcd}, Long-tailed NCD~\cite{zhang2023novel},\\ Long-tailed GCD~\cite{li2023generalized}, BYOP~\cite{yang2023bootstrap}, DA-GCD~\cite{bai2023towards}} \\
                \cline{3-4}
                & & Multimodal Discovery & \makecell{SCD~\cite{han2023s}, CLIP-GCD~\cite{ouldnoughi2023clip}, TextGCD~\cite{zheng2024textual},\\ GET~\cite{wang2024get}, FineR~\cite{liu2024democratizing}, CPT~\cite{yang2025consistent}} \\
                \cline{2-4}
                \hline
                \multirow{20}{*}{\makecell{Continual Learning}} &\multirow{3}{*}{\makecell{Regularization \\
Method}}
                & Explicit Weight Constraint & \makecell{EWC \cite{Kirkpatrick2017OvercomingCF}, SI \cite{Zenke2017ContinualLT}, MAS \cite{aljundi2018memory}, RWalk \cite{chaudhry2018riemannian}, VCL \cite{nguyen2018variational}, \\ UCL \cite{ahn2019uncertainty}, KCL \cite{derakhshani2021kernel}, AGS-CL \cite{jung2020continual}, OWM \cite{zeng2019continual}, OGD \cite{farajtabar2020orthogonal},\\ AOP \cite{guo2022adaptive}, GPM \cite{sahagradient},  Adam-NSCL \cite{wang2021training}, NCL \cite{kao2021natural}, CAF~\cite{wang2023incorporating}} \\
                \cline{3-4}
                & & Implicit Knowledge Distillation & \makecell{LwF~\cite{hinton2015distilling}, EBLL~\cite{rannen2017encoder}, LwM~\cite{DharSPWC19},  DMC~\cite{zhang2020class}, M2KD~\cite{zhou2019m2kd},\\ MUC~\cite{liu2020more}, CCIL~\cite{zhu2021calibration}, UCIR~\cite{Hou2019LearningAU}, PODNet \cite{Douillard2020SmallTaskIL}, DER \cite{buzzega2020dark},\\ DER++ \cite{buzzega2020dark}, GeoDL \cite{simon2021learning}, DDE \cite{hu2021distilling}, FOSTER \cite{wang2022foster} }\\
                \cline{2-4}
                & \multirow{3}{*}{\makecell{Data Replay\\ Methods}} & Imbalance Calibration & \makecell{EEIL~\cite{castro2018end}, UCIR~\cite{Hou2019LearningAU}, GDumb~\cite{prabhu2020greedy}, iCaRL~\cite{Rebuffi2017iCaRLIC}, BiC~\cite{Wu2019LargeSI}, \\ WA~\cite{ZhaoXGZX20}, ScaIL~\cite{belouadah2020scail}, IL2M~\cite{belouadah2019il2m}, SS-IL~\cite{ahn2021ss}, ItO \cite{zhu2023imitating}} \\
                \cline{3-4}
                & & Generative Data Replay & \makecell{ DGR~\cite{shin2017continual}, MeRGAN~\cite{wu2018memory}, FearNet~\cite{kemker2018fearnet}, DMC~\cite{zhang2020class}, \\ ABD~\cite{smith2021always}, ILCAN~\cite{xiang2019incremental}, DGM~\cite{liu2020generative}, \\ DiffClass~\cite{meng2024diffclass}, DDGR~\cite{gao2023ddgr}, SDDGR~\cite{kim2024sddgr}} \\
                \cline{2-4}
                & \multirow{5}{*}{\makecell{Feature Replay\\ Methods}} & Real / Generative Feature & \makecell{REMIND~\cite{hayes2020remind}, LR~\cite{pellegrini2020latent}, GFR~\cite{liu2020generative}, \\ BI-R~\cite{van2020brain}, FAdapt \cite{iscen2020memory}} \\ 
                \cline{3-4}
                & & Prototype Replay & \makecell{SDC \cite{yu2020semantic}, PASS~\cite{zhu2021prototype}, PASS++~\cite{zhu2024pass++}, SSRE \cite{zhu2022self}, IL2A~\cite{zhu2021class}, \\Fusion~\cite{toldo2022bring}, FeTrIL~\cite{petit2023fetril}, FeCAM~\cite{goswami2023fecam}, SOPE \cite{zhu2023self}, NAPA-VQ \cite{malepathirana2023napa}, \\ADC \cite{goswami2024resurrecting}, LDC~\cite{gomez2024exemplar}, DPCR \cite{he2025semantic}, EFC \cite{magistri2024elastic}, EFC++ \cite{magistri2025efc++}, \\AdaGauss \cite{rypesc2024task}, RanPAC \cite{mcdonnell2023ranpac}, SDCLoRA \cite{wu2025navigating}, PRL \cite{shi2024prospective}, \\Semi-IPC~\cite{liu2024towards}, FCS~\cite{li2024fcs}} \\
                \cline{2-4}
                & \multirow{6}{*}{\makecell{Model Expansion\\ Methods}} & Backbone Expansion & \makecell{P\&C \cite{schwarz2018progress}, AANets~\cite{liu2021adaptive}, DER~\cite{yan2021dynamically}, Piggyback~\cite{mallya2018piggyback}, \\HAT~\cite{serra2018overcoming}, PackNet~\cite{mallya2018packnet}, DyTox~\cite{douillard2022dytox}} \\
                \cline{3-4}
                & & \makecell{Parameter Efficient\\ Expansion} & \makecell{L2P~\cite{wang2022learning}, CODA-Prompt~\cite{smith2022coda}, DualPrompt~\cite{wang2022dualprompt}, SDCLoRA \cite{wu2025navigating}, \\PILoRA \cite{guo2024pilora}, SD-LoRA \cite{wu2025sd}, Online-LoRA \cite{wei2025online}, LAE \cite{gao2023unified}, \\InfLoRA \cite{liang2024inflora}, HiDe-Prompt \cite{wang2023hierarchical}, C-LoRA \cite{zhang2025c},C-CLIP \cite{liu2025c}, \\TSVD \cite{peng2025tsvd}, CAPrompt \cite{li2025caprompt}, DESIRE \cite{guo2024desire}, FM-LoRA \cite{yu2025fm}, \\ LGCL~\cite{khan2023introducing}, PGP~\cite{qiao2023prompt}, EvoPrompt~\cite{kurniawan2024evolving}, RCS-Prompt~\cite{yang2024rcs}, \\MISA \cite{zhiqi2025advancing}, LORA-DRS~\cite{liu2025lora},  EASE~\cite{zhou2024expandable}, NoRGa~\cite{le2024mixture}, \\ SAFE~\cite{zhaosafe}, VQ-Prompt~\cite{jiaovector}, VPT-NSP~\cite{luvisual}, RAPF~\cite{huang2024class}, \\ C-ADA~\cite{gao2024beyond}, PromptFusion~\cite{chen2024promptfusion}
                } \\
                \cline{2-4}
                & \multirow{4}{*}{\makecell{Extended Settings}}
                & \makecell{Unsupervised\\ Continual Learning} & \makecell{FRoST~\cite{roy2022class}, NCDwF~\cite{joseph2022novel}, GM~\cite{zhang2022grow}, IGCD~\cite{Zhao_2023_ICCV_Incremental}, PA-GCD~\cite{Kim_2023_ICCV}, \\ PromptCCD \cite{cendra2024promptccd}, Happy~\cite{NEURIPS2024_5ae0f7cf}}\\
                \cline{3-4}
                & & \makecell{Multimodel\\ Large Language Model} & \makecell{O-LoRA \cite{wang2023orthogonal}, HiDe-LLaVA \cite{guo2025hide}, ModalPrompt \cite{zeng2024modalprompt}, \\ MOELoRA \cite{chen2024coin}, Continual-LLaVA \cite{cao2024continual}, DISCO \cite{guo2025federated}, LLaCA \cite{qiao2024llaca}, \\ LLaVA-CMoE \cite{zhao2025llava}, LOIRE~\cite{hanloire}, FV~\cite{jiangunlocking}, Adapt-$\infty$~\cite{maharanaadapt}, \\ SAPT~\cite{zhao2024sapt}, ProgPrompts \cite{razdaibiedina2023progressive}, DAS~\cite{ke2023continual}
                }\\
                \hline
                \end{tabular}}
\end{center}
\end{table*}

However, as illustrated in Fig. \ref{Figure:intro}, real-world applications often involve dynamic and open environments, where unexpected situations inevitably arise and instances belonging to unknown classes ($y \notin \mathcal{Y}$) may appear \cite{zhou2022open, masana2022class}. For example, in non-stationary environments, a self-driving car may encounter novel objects that have never been learned before; myriad novel categories would emerge continually in web usage and face recognition systems. 
The closed-world assumption can be problematic in such situations. Firstly, models are overconfident and predict unknowns as training classes without hesitation \cite{hendrycks2016baseline, zhu2022learning, zhang2023survey}, which can cause various harms from financial loss to injury and death. Secondly, models fail to extrapolate to novel classes by discovering and clustering them \cite{han2021autonovel} from various unknowns. Thirdly, learning new stream data leads to catastrophic forgetting of previous knowledge \cite{mccloskey1989catastrophic}.  
To learn in such an endless variety of ever-changing scenarios, we need open-world learning to overcome these limitations by accommodating the dynamic and uncertain nature of real-world data. In this paradigm, a model is equipped to identify and reject inputs that deviate from training classes to keep safe, and then discovers meaningful new classes from unknowns and continually learns them to accumulate knowledge without re-training the whole model from scratch. 

The general life cycle of an open-world learning (OWL) paradigm is illustrated in Fig.~\ref{figure-2}, mainly consisting of three key steps. The first step is unknown rejection, which requires the model to recognize test instances that belong to seen classes while also being able to detect or reject misclassified and unknown instances that do not belong to the training classes based on reliable confidence estimation \cite{zhu2022learning, zhu2023revisiting}. The second step is novel class discovery~\cite{han2021autonovel}, which clusters the collected unknown samples and identifies meaningful categories in the buffer automatically based on the knowledge learned in the past. Finally, when the discovered classes have sufficient data, the system must extend the original multi-class classifier to incorporate new classes with continual learning techniques, avoiding retraining from scratch or catastrophic forgetting of previously learned knowledge \cite{van2022three, verwimp2023continual, wang2023comprehensive}.
By integrating unknown rejection, novel class discovery, and continual learning, the system is able to scale and adapt to an ever-evolving environment. In other words, the model can be aware of what it does not know and learns interactively after deployment (on the job) in the open world like humans.

OWL has garnered significant attention and exploration in recent years, resulting in the emergence of numerous methodologies in this research field. The exploration of OWL traces its conceptual roots to early investigations into open-set recognition and continual learning, with foundational works around the 2010s emphasizing statistical and shallow models to detect unknowns or adapt to incremental data \cite{zhou2022open, mccloskey1989catastrophic}. An example is the introduction of open-set recognition frameworks \cite{scheirer2014probability}, which aimed to reject unseen categories by quantifying uncertainty in decision boundaries, though these methods often struggled with high-dimensional real-world data. The advent of deep learning \cite{goodfellow2016deep, pouyanfar2018survey} marked a paradigm shift, as neural networks demonstrated superior capability in representation learning. Notably, Hendrycks \etal~\cite{hendrycks2016baseline} established a baseline using deep networks to detect out-of-distribution samples, highlighting the role of softmax confidence scores in identifying unknowns. This era also witnessed growing attention to catastrophic forgetting \cite{mccloskey1989catastrophic}, spurring research into continual learning techniques to preserve historical knowledge while integrating new information. In recent years, pre-training models have been leveraged to enhance the reliability and adaptation of AI systems in open environments \cite{zhu2023revisiting, masana2022class}. These advancements underscore a shift toward end-to-end systems capable of self-awareness and autonomous knowledge expansion. Looking ahead, OWL remains a vibrant yet challenging frontier. Future research is to explore OWL with foundation models, a broader spectrum of data modalities, and neuromorphic mechanisms to achieve human-like adaptability in ever-evolving environments.

Although prior surveys~\cite{zhou2022open, parmar2023open,zhang2023survey} have reviewed some aspects of OWL, significant gaps persist in terms of timeliness, taxonomy and comprehensiveness, \ie, they are either relatively early reviews lacking discussions on the up-to-date advances, or limited to only a few aspects of OWL. Specifically, Zhou~\cite{zhou2022open} briefly introduced the concept of emerging new classes, decremental and incremental features, as well as changing data distributions in OWL. Parmar \etal~\cite{parmar2023open} mainly focused on open-set classification, and Zhang \etal~\cite{zhang2023survey} offered a discussion of scenarios where the model needs to learn to reject. However, these works focus mainly on early research and a specific aspect of OWL, while important methods developed in the last five years, which have significantly shaped the field, have yet to be included. 
To overcome the aforementioned limitations and fill the gap in the literature of OWL, we present this comprehensive survey with a broader spectrum of learning tasks of OWL, including all three steps in Fig.~\ref{figure-2} (c). For each task, we further provide a more detailed sub-taxonomy with up-to-date research advances.
We undertake a comprehensive review of over four hundred articles, introducing a taxonomy delineating unknown rejection, novel class discovery, and continual learning. As shown in Table~\ref{tab:summary}, for each sub-topic, we divide existing methods into overarching categories and further into several subcategories. 
The principles and limitations of current methods and the relationship among them are discussed. Besides, existing theoretical results are also presented. 
We also summarize the commonly used datasets and evaluation metrics, and systematically evaluate different methods across various benchmarks. Furthermore, the possible challenges, research gaps, and potential research directions for the future development of open-world machine learning are presented. This survey updates the field’s narrative, being helpful for researchers to apply this new learning paradigm in their own domains, and also calls for building human-like, truly intelligent systems.

The rest of this paper is organized as follows: Section \ref{sec:background} provides the basic problem formulation, notations, several canonical use cases, and an overview of open-world machine learning. Section~\ref{sec:unknown-rejection} presents unknown rejection, which requires the model to detect test-time unknown classes outside the score of trained classes. Section~\ref{sec:novel-discovery} further introduces novel class discovery, including novel category discovery and its generalized setting, which requires further clustering of the new classes. Section~\ref{sec:class-incremental-learning} elaborates on continual learning. Finally, Section~\ref{sec:future} outlines future directions and Section~\ref{sec:conclusion} concludes this paper.

\begin{figure*}[t]
	\begin{center}
		\centerline{\includegraphics[width=0.95\textwidth]{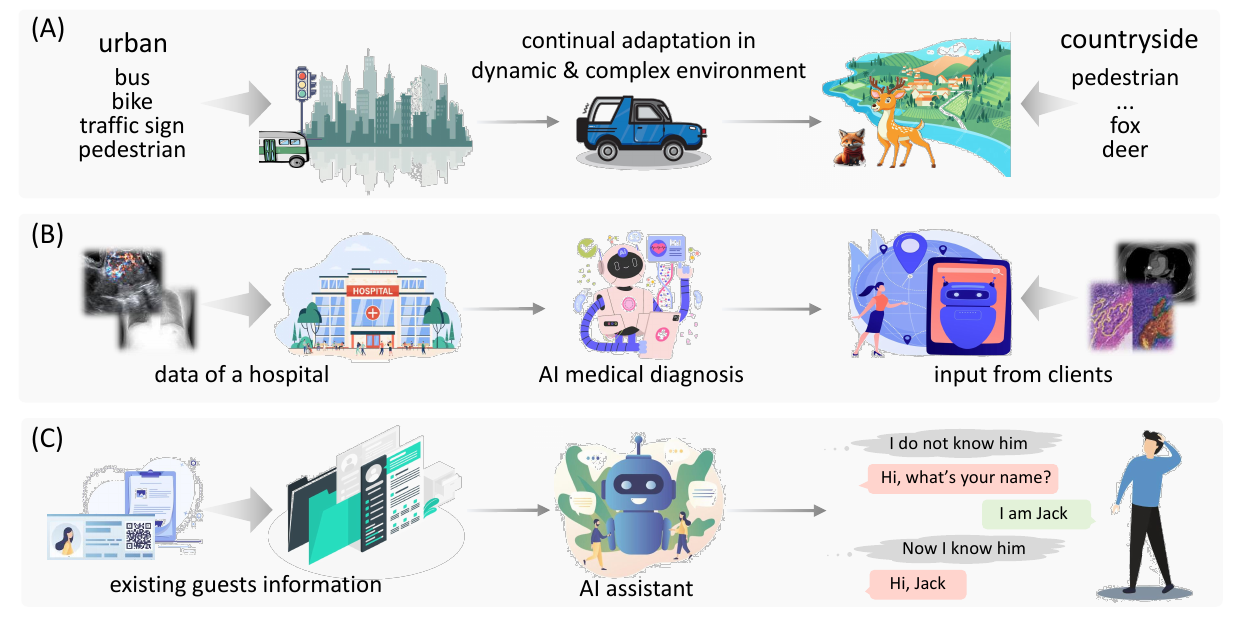}}
		\vskip -0.1in
		\caption{Example applications of open-world machine learning systems. (A) Autonomous driving can leverage open-world learning to handle unknown objects and evolving environments, enabling safe navigation, adaptation to dynamic environments. (B) Medical diagnosis can encounter diverse input distributions from clients, and open-world learning enables more reliable clinical decisions and responsiveness. (C) AI chatbot continually learns from user-specific information and improves the dialogue behavior over time.}
		\label{figure-3}
	\end{center}
	\vskip -0.2in
\end{figure*}
\section{Background and Overview}
\label{sec:background}

\subsection{Basic Notations}
Here, we give a holistic problem formulation and notations of open-world machine learning, comprising three tasks: unknown rejection, novel class discovery and continual learning. Let $\mathcal{D}_\text{train}=\{(\mathbf{x}_i,y_i)\}_{i=1}^N\subset \mathcal{X}\times \mathcal{Y}_\text{train}$ denote the training dataset, and $\mathcal{D}_\text{test}=\{(\mathbf{x}_i,y_i)\}_{i=1}^M\subset \mathcal{X}\times \mathcal{Y}_\text{test}$ denote the test dataset. By default, the training data only contains labeled data from old and known classes, \ie, $\mathcal{C}_{\text{old}}=\mathcal{Y}_\text{train}$. We represent a deep neural network (DNN) based model $f$ with two components: a feature extractor $h$ and a unified classifier $g$, \ie, $f = g \circ h$. In the canonical closed-world setting, the test dataset shares categories with training data with no additional classes, \ie, $\mathcal{Y}_\text{test}=\mathcal{Y}_\text{train}$. By contrast, in the open world, samples from new classes emerge, \ie, $\mathcal{Y}_\text{train}\neq \mathcal{Y}_\text{test}$, we further express it as $\mathcal{Y}_\text{train}\subset \mathcal{Y}_\text{test}$. In this paper, we not only consider how to classify the seen and known classes, but also focus on how to handle samples from new classes $\mathcal{C}_{\text{new}}=\mathcal{Y}_\text{test}\setminus \mathcal{Y}_\text{train}$, which is divided into three progressive ways: unknown rejection, novel class discovery and continual learning. In unknown rejection, models only need to reject unknowns, while in novel class discovery and continual learning, models are further required to classify/cluster samples from new classes. Considering the clustering objective, in novel class discovery, $\mathcal{D}_\text{train}$ could also contain some unlabeled data from $\mathcal{C}_{\text{new}}$. While in continual learning, the training scheme could be divided into several stages, models are trained on streaming data, and we add a superscript $t$ to the dataset $\mathcal{D}_\text{train}^t$ and $\mathcal{D}_\text{test}^t$ to indicate each continua step. Considering the differences among the three tasks, we will also give task-specific notations for the three tasks when discussing them.

\subsection{Canonical Use Cases}
In this section, we discuss several canonical applications of open-world machine learning and illustrate them in Fig. \ref{figure-3}.

\textbf{\emph{1) Autonomous Driving.}} 
Self-driving cars in real-world scenarios can encounter a nearly infinite number of unexpected and novel objects \cite{suenderhauf2019probabilistic}, and open-world learning can provide them with the adaptability needed to navigate safely and effectively. Specifically, when encountering a novel object on the road, \eg, a deer or even a hole on the road that does not appear in the training class set, the car should stop or slow down. This involves leveraging an unknown rejection algorithm that provides reliable confidence scores. If the confidence score is lower than a threshold, the system should raise an alarm so that the car takes a predefined safety action. Then, novel class discovery techniques are used to create the class identity for the unknown, new object. Finally, based on continual learning methods, the car can continue to learn from a single or few samples of the object, enabling autonomous systems to quickly adapt to unexpected new environments without forgetting the skills already mastered. In conclusion, open-world learning allows the vehicle to learn from dynamic real-world environments and adapt its decision-making process, improving its driving behavior over time.

\textbf{\emph{2) Medical Diagnosis.}} Real-time medical applications require the AI system to address many heterogeneous problems that involve multiple, complex tasks. For example, a pulmonary model trained on existing lung data would not be able to reliably predict newly emerged pulmonary diseases like COVID-19, and the medical diagnostic tools can easily make mistakes \cite{park2021reliable}. In such cases, the model should be equipped with the unknown rejection ability to output reliable and human-interpretable confidence scores along with its predictions, which helps doctors or participants to make safe decisions. For example, the input should be handed over to human doctors when the confidence of a disease diagnosis is low. When encountering more and more unknowns, the medical diagnosis system is supposed to continually learn from those unknowns or more data to improve its performance progressively \cite{lee2020clinical, vokinger2021continual}. Since manual image grading is time-consuming, novel class discovery methods would largely accelerate the processing of labeling new data. The above adaptive capability ensures that medical diagnosis systems can learn from on-the-job experiences and apply this knowledge to similar situations in the future, which is very similar to the ways that human clinicians learn. In conclusion, OWL enables medical diagnoses to make more reliable clinical decisions and be more responsive to evolving environments.

\textbf{\emph{3) AI Chatbot.}} With the launch of large language models like ChatGPT \cite{OpenAI2023GPT4TR}, AI Chatbots have become widely used in many scenarios to help with daily tasks for users. In practice, it is critical for a Chatbot to know what it doesn't know. Specifically, if the Chatbot does not know the answer, it is better to reject to response rather than provide incorrect information that may cause very serious mistakes in fields like clinical and legal matters \cite{zhuo2023exploring, whitehead2022reliable}. One way to avoid providing misleading information is to abstain from making a prediction based on unknown rejection methods. Further, if the user provides the correct answer or more data is available, the Chatbot should improve its conversational skills continually without forgetting previously learned knowledge. A typical example is the hotel guest-greeting bot \cite{chen2018lifelong}, where a Chatbot greets known hotel guests, detects new guests, and then continually learns them by asking for their names and taking photos. As can be seen, open-world learning makes the AI Chatbot more knowledgeable and powerful over time.

\subsection{Overall Challenges of Open-world Learning}

As illustrated in Fig.~\ref{figure-2} and Fig.~\ref{figure-3}, open-world learning involves performing unknown rejection, novel class discovery, and continual learning sequentially and periodically. The core challenge is enabling the above process to automatically proceed through the interactions between the model and the open environments without relying on human engineers \cite{liu2023ai}. Unfortunately, under the closed-world assumption \cite{zhang2020towards}, a model is overconfident and can hardly be aware of the unknown. Specifically, from the perspective of representation learning, the model is trained only on the current dataset with data-driven optimization, the learned representations are task-specific and less generic; from the perspective of classifier learning, current discriminative classifiers leave little room for the unknown, making it hard to characterize, discover and adapt to novelty. Consequently, examples from unknown classes would be mapped to the region of known classes, leading to catastrophic forgetting of previous knowledge in the latter continual learning process.

\section{Unknown Rejection}
\label{sec:unknown-rejection}

\subsection{Problem Formulation and Key Challenges}
\textbf{\emph{1) Problem Formulation.}} Unknown rejection is the first step towards open-world machine learning, which is also a fundamental ability of the classifier in the open-world. Considering that a training set comprises of $K$ classes, \ie, $\mathcal{D}_\text{train}=\mathcal{D}_\text{in}=\{(\mathbf{x}_i,y_i)\}_{i=1}^N\subset\mathcal{X}\times \mathcal{Y}_\text{in}$, where $\mathcal{Y}_\text{in}=\{1,2,\cdots,K\}$. Once trained on $\mathcal{D}_\text{in}$, models are supposed to reject unknown samples from classes outside of $\mathcal{Y}_\text{in}$, rather than classifying them irresponsibly into one of $K$ categories. The philosophy behind unknown rejection is that when encountering unfamiliar knowledge, it is essential to bravely acknowledge the limitations rather than feign understanding and give arbitrary answers.

\begin{figure*}[t]
  \begin{center}
\centerline{\includegraphics[width=\textwidth]{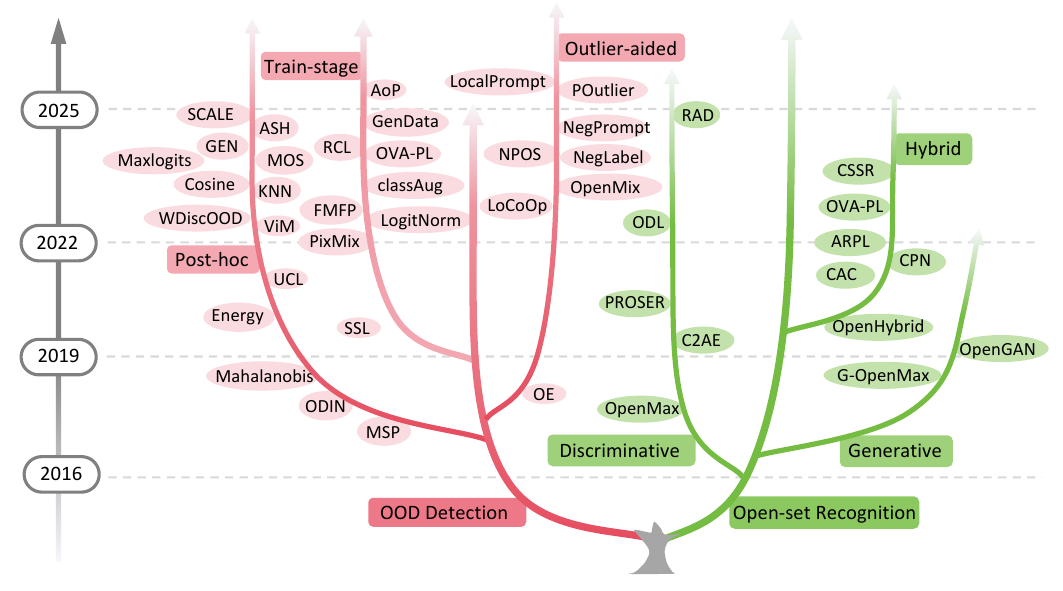}}
   \caption{The evolutionary tree of unknown rejection methods.}
   \label{treefig_ur}
 \end{center}
 \vskip -0.2in
\end{figure*}

\begin{figure*}[t]
	\centering
	\includegraphics[width=0.97\textwidth]{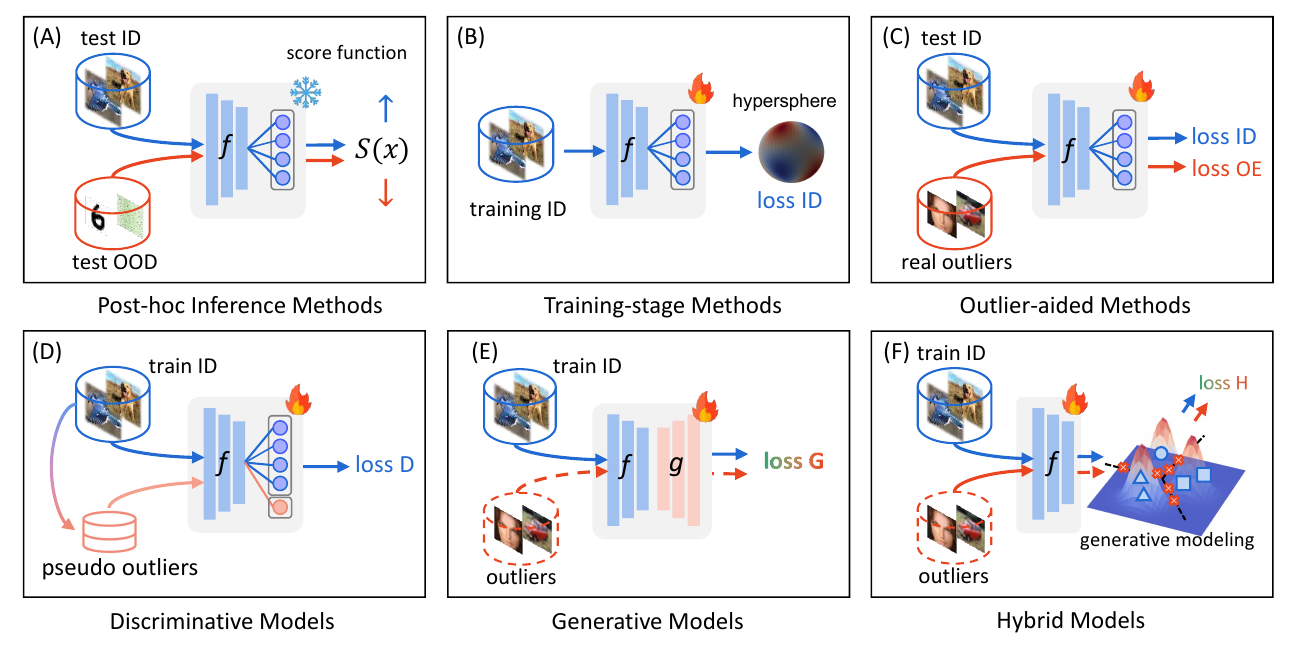}
	\vskip -0.15in
	\caption{Illustrations of different types of methods of OOD detection (first row) and Open-Set Recognition (OSR, second row). In OOD detection, methods are divided into three types, \ie, (a), (b) and (c), according to training and inference strategies. For OSR, methods are also divided into three types, \ie, (d), (e) and (f), regarding modeling perspectives in OSR. In OSR, ``loss D'', ``loss G'' and ``loss H'' denote loss functions for discriminative, generative and hybrid models, respectively. Here, $f$ and $g$ denote the feature extractor and generator, respectively.}
	\label{fig:osr-ood}
\end{figure*}

Many efforts have been made to enhance the unknown rejection ability of machine learning systems. There are multiple research areas related to unknown rejection, such as anomaly detection~\cite{chandola2009anomaly,pang2021deep}, out-of-distribution (OOD) detection~\cite{yang2021generalized,salehi2022a}, and open-set recognition~\cite{scheirer2012toward,geng2020recent} (OSR). Among them, anomaly detection is widely used in early work, while OOD detection and OSR are more often used in recent studies. The differences between OOD detection and OSR lie in two aspects. First, in OOD detection, the semantic distance between the OOD data and in-distribution (ID) data is relatively larger, \eg, the OOD dataset is usually completely unrelated. While in OSR, a subset of classes from a dataset is viewed as ID, and other classes from the same dataset are viewed as OOD data. Second, OOD detection mainly concentrates on distinguishing OOD samples from ID samples, whereas OSR also evaluates closed-world classification performance on known classes. We overview the recent advances in OOD detection and OSR in the remaining part of this section, and different types of methods are summarized in Fig.~\ref{treefig_ur} and illustrated in Fig.~\ref{fig:osr-ood}.

\textbf{\emph{2) Key Challenges.}}
The inherent challenge of unknown rejection is that OOD samples are agnostic during training. Even though some methods~\cite{hendrycks2018deep,zhang2023mixture} resort to auxiliary outlier OOD samples, it is impossible to encompass all OOD distributions exhaustively. As a consequence, models need to reserve some open space for potential unknown samples to reduce the open space risk~\cite{scheirer2012toward,geng2020recent}. Secondly, when the semantic similarity~\cite{vaze2022openset} between OOD and ID samples is minimal, their separability decreases, resulting in a decline in rejection capability. Such samples are referred to as \emph{near-ood} samples in the literature~\cite{yang2022openood}. Furthermore, some issues could impact the rejection of unknown samples, including covariate shifts~\cite{bai2023feed} and spurious correlations~\cite{ming2022spurious}, for example, ID samples from another distribution domain might be mistakenly detected as OOD samples, and spurious correlations learned from training data could mislead the models to focus on irrelevant features and areas.
In this section, we will introduce two types of tasks related to rejection, \ie, OOD detection (Sec.~\ref{subsec:ood}) and Open-set Recognition (Sec.~\ref{subsec:osr}), respectively, as well as their representative methods and comparative results. Then, the theoretical advance of unknown rejection is summarized (Sec.~\ref{subsec:reject-analysis}).
Finally, evaluation results of unknown rejection are presented (Sec.~\ref{subsec:evaluation-ur}).

\subsection{Out-of-distribution Detection}
\label{subsec:ood}

OOD detection~\cite{yang2021generalized,salehi2022a,ma2025towards} focuses on whether a test example is outside the distribution of the training data and aims to improve the separability between ID and OOD samples. Technically, the common practice is to assign each sample a score function $S(\boldsymbol{x})$, indicating the \emph{otherness} of each sample. If the score is above a pre-defined threshold $\delta$, then the sample is recognized as ID, otherwise it is detected as OOD and rejected. This formula could be written in Eq.~\eqref{eq:ood-formula}. OOD detection methods can be categorized into three classes~\cite{yang2022openood,zhang2023openood} based on different training and inference techniques.
\begin{equation}
    g(\boldsymbol{x})=\left\{
        \begin{aligned}
        & \text{in-distribution}, \quad & \text{if}\ S(\boldsymbol{x})\geq \delta, \\
        & \text{out-of-distribution}, \quad & \text{if}\ S(\boldsymbol{x})< \delta.
        \end{aligned}
        \right.
    \label{eq:ood-formula}
\end{equation}

\textbf{\emph{1) Post-hoc Inference Methods.}}
One natural and simple way is to design appropriate score function $S(\boldsymbol{x})$ that has the most significant separability between ID and OOD samples. Post-hoc methods are simply implemented on the trained models, without interfering with the training phase, as a result, they are orthogonal to the training-stage methods discussed below. Many characteristics of the trained models could be designed as score functions, such as posterior probabilities of the classification head, activations and dynamics in the feature space. Early works design $S(\boldsymbol{x})$ based on probabilities, \ie, $p(c|\boldsymbol{x})=\frac{\exp(\boldsymbol{z}_c(x))}{\sum_{j=1}^K\exp(\boldsymbol{z}_j(x))}$ where $\boldsymbol{z}_c(x)$ denotes the logits of sample $\boldsymbol{x}$. Hendrycks \etal~\cite{hendrycks2016baseline} proposed a simple score function, namely maximum softmax probability (MSP), \ie, $S(\boldsymbol{x})=\max_c p(c|\boldsymbol{x})$, which is regarded as a baseline method in the literature. MSP could be enhanced by input perturbation, which aims to increase the softmax score for the true label and temperature scaling~\cite{liang2018enhancing}. Apart from softmax probabilities, energy score~\cite{liu2020energy} derived from energy-based models~\cite{lecun2006tutorial} is less susceptible to the overconfidence issue of classifiers, achieving better OOD detection abilities. Similar to probabilities, logits also serve as the belongingness of each class without the normalization process in MSP, and the maximum logit score (Maxlogits)~\cite{hendrycks2022scaling} is reported better than the MSP score. Instead of the probability distribution, characteristics in the feature space could also be exploited for OOD detection. Lee \etal~\cite{lee2018simple} designed the Mahalanobis distance-based confidence score by modeling layer-wise Gaussian distribution of features, which could be substituted by flow-based~\cite{kingma2018glow} generative modeling for more flexible expressivity~\cite{zisselman2020deep}. These methods still impose certain assumptions on the distribution of the feature spaces, which limits the performance to some extent. By contrast, non-parametric nearest-neighbor distance (KNN)~\cite{sun2022out} is more general and flexible without relying on any distributional assumptions. Activations in the penultimate layer~\cite{sun2021react} could also be considered. The methods above design $S(\mathbf{x})$ from only one of the two aspects. We could also combine the above two perspectives~\cite{wang2022vim}, namely posterior probabilities and feature embeddings, for a more remarkable separation between ID and OOD samples. Virtual-logit matching (ViM)~\cite{wang2022vim} combines the class-agnostic information in the feature space and the class-dependent information in logits. Recently, activation shaping has been demonstrated to be quite effective for OOD detection. For example, ASH~\cite{djurisic2022extremely} enhances OOD detection performance remarkably by pruning and lightly adjusting a large portion of late-layer activations during inference. Similarly, by scaling network activations, SCALE~\cite{xu2023scaling} achieves state-of-the-art performance without compromising in-distribution accuracy.

\textbf{\emph{2) Training-stage Methods.}}
Training-based methods directly explore the training of models with inherent capabilities to discriminate between ID and OOD samples. This line of works are complementary to the post-hoc methods dedicated to designing OOD scores, they could be combined for better OOD detection performance. In general, advanced training tricks, including self-supervised learning~\cite{hendrycks2019using} and normalization~\cite{wei2022mitigating}, provide outstanding OOD detectors. Self-supervised learning, like rotation prediction~\cite{gidaris2018unsupervised} and contrastive learning~\cite{chen2020simple}, could help improve robustness and uncertainty estimation, which also greatly boosts OOD detection~\cite{hendrycks2019using,tack2020csi} with alleviated overconfidence issues. Another technique, namely logit normalization~\cite{wei2022mitigating}, helps to mitigate the overconfidence issue for both ID and OOD samples. Models could also be encouraged for greater inter-class separation and intra-class compactness~\cite{ming2023how} in the feature space, which leaves more room for open space and helps separate ID and OOD samples. Recently, flat minima has been demonstrated to be a good optimization objective for the detection of misclassified examples from known classes and OOD examples from unknown classes~\cite{zhu2023revisiting}. Recently, Chen \etal~\cite{cheng2024breaking} revealed that training with large-scale synthetic images from generative models can effectively eliminate overfitting in OOD detection, enabling small models like ResNet-18 \cite{he2016deep} trained on generated data to rival ViT \cite{dosovitskiy2021an} pretrained on massive real datasets. When scaling OOD detection to large-scale datasets, a key insight is to decompose the large semantic space into many small groups and train models with group-based learning~\cite{huang2021mos}. TrustLoRA \cite{zhu2025trustlora} introduces the low-rank adaptation technique into unknown rejection and unifies rejection under both covariate and semantic shifts via reliability consolidation.

\textbf{\emph{3) Outlier-aided Methods.}}
It is also intuitive to directly train models on auxiliary OOD samples $\mathcal{D}_\text{out}$. By maximizing their uncertainty, the model tends to assign lower confidence to OOD samples, which helps increase the separability between OOD and ID samples. Outlier exposure~\cite{hendrycks2018deep} (OE) is a pioneering work to leverage auxiliary outliers during training. Specifically, OE proposes to maximize the uncertainty on the outliers. The loss function of OE is:
\begin{equation}
    \mathcal{L}=\mathbb{E}_{(\boldsymbol{x},y)\in\mathcal{D}_\text{in}}\mathcal{L}_{ce}(y,f(\boldsymbol{x})+\lambda\mathbb{E}_{\boldsymbol{x}\in \mathcal{D}_\text{out}} \mathcal{L}_{ce}(\mathcal{U},f(\boldsymbol{x})),
    \label{eq:oe}
\end{equation}
where $f(\boldsymbol{x})$ is the predictive probability, $\mathcal{L}_{ce}(\cdot,\cdot)$ is the cross-entropy loss, and $\mathcal{U}$ is the uniform distribution. Practically, OE aims to minimize the cross-entropy between the uniform distribution and the model's predictions to encourage uncertain predictions on these outlier samples. The authors~\cite{hendrycks2018deep} found that once learned on $\mathcal{D}_\text{out}$, the model could generalize to detect unseen OOD samples and anomalies and achieve remarkable OOD detection performance. 
Zhu \etal~\cite{zhu2023openmix} revealed that popular OOD detection methods like OE often make it harder to detect misclassified examples from known classes, and proposed a unified misclassification and OOD detection approach named OpenMix. Besides, taking the spirit of MixUp~\cite{zhang2018mixup}, mixing ID training samples with auxiliary outliers~\cite{zhang2023mixture} could expand the coverage of different OOD granularities and enhance the performance of OE.  Virtual Outlier
Synthesis (VOS) \cite{du2022vos} models the penultimate feature embeddings of each class as a multivariate Gaussian, then samples virtual outliers from the low‐likelihood regions of these class‐conditional distributions and uses them to tighten the ID/OOD decision boundary. Ma \etal~\cite{ma2025towards} extend the setting of OOD detection to the data-efficient scenarios~\cite{zhao2021dataset,yu2023dataset,lei2023comprehensive}, and propose to synthesize pseudo-outliers using ID data, which could also improve the detection performance without the requirement of real outliers. Recent studies \cite{miyai2023locoop, li2024learning, jiang2024negative, zeng2025local} focus on few-shot OOD detection of vision–language models with pseudo-outliers mining. For example, LoCoOp \cite{miyai2023locoop} leverages local image features as negative OOD examples during prompt training to regularize and purify the context embeddings, thereby removing ID-irrelevant nuisances and enhancing ID/OOD separation.
LocalPrompt \cite{zeng2025local} uses prompt guided negative augmentation to inject local outlier cues to improve the performance of rejecting unknown rejection.

\subsection{Open-Set Recognition}
\label{subsec:osr}

OSR~\cite{scheirer2012toward,geng2020recent} aims to recognize known in-distribution classes while detecting unknown OOD classes at the same time. In the literature of OSR, ID and OOD are conventionally drawn from different class sets within the same dataset. Early OSR works are mainly based on support vector machine~\cite{scheirer2012toward}, nearest class mean classifiers~\cite{mensink2013distance}, and extreme value machine~\cite{zhang2016sparse}. Those traditional OSR approaches have been well reviewed and discussed \cite{geng2020recent}. In recent years, various deep learning-based OSR methods have been developed, and we mainly focus on them in this paper. The methods could be divided into three types from the modeling perspective: discriminative, generative and hybrid models.

\textbf{\emph{1) Discriminative Models.}}
Discriminative models perform relatively well in classification tasks, it is natural to employ them to handle OSR, \ie, classify among known classes and reject unknown classes. One direct and intuitive method is to leave room for the modeling of unknown classes. A pioneer work, OpenMax~\cite{bendale2016towards} recalibrates the logits of top classes from known to assign the probability to the appended class \emph{zero} representing unknown classes. OpenMax~\cite{bendale2016towards} resorts to the extreme value theorem and utilizes the Weibull distribution to compute the re-calibration weights. Dummy classifiers~\cite{zhou2021learning} for unknown samples could mimic the novel classes as data placeholders and augment the closed-set classifier with classifier placeholders. From another perspective, similar to OOD detection, models could also be trained on both known and auxiliary unknown classes to behave differently on knowns and unknowns and improve the ability to distinguish between them. Dhamija \etal~\cite{dhamija2018reducing} leveraged samples from background classes as unknown samples and proposed two novel loss functions to make ID and OOD well-separated in both probability space and feature space. DTL~\cite{perera2019deep} proposes membership loss to substitute cross-entropy loss for class-wise probability modeling, and also relies on outlier data to train a parallel network for providing global negative filtering. The methods discussed above mainly employ the output of classifiers as a proxy to discern ID and OOD samples. Besides, autoencoders~\cite{rumelhart1985learning,Goodfellow-et-al-2016} are proven to be good indicators for OSR. The fundamental principle is that models are trained on known samples to minimize reconstruction errors and will produce relatively higher reconstruction errors on unknown samples. C2AE~\cite{oza2019c2ae} divides OSR into closed-set classification and open-set identification, and the test sample is rejected as unknown if the reconstruction error is below the threshold. 
The encoder is first trained on known training samples to learn the classifier. The second stage is conditional decoder training to reconstruct inputs conditioned on class identity. C2AE~\cite{oza2019c2ae} employs EVT to obtain the threshold for unknown identification. The test sample is rejected as unknown if the reconstruction error is below the threshold. Furthermore, maintaining hierarchical reconstructions~\cite{yoshihashi2019classification} with various levels of abstraction brings about more stable results. 

The ability of a classifier to reject unknowns has been verified as highly correlated with its closed-set classification accuracy~\cite{vaze2022openset}. As a result, when applying some training tricks to the classifier for better closed-set classification, \eg, longer training, RandAugment~\cite{cubuk2020randaugment}, learning rate warmup, and label smoothing~\cite{szegedy2016rethinking}, the classifier consequently performs better in OSR. The authors~\cite{vaze2022openset} further found that applying baseline methods like MSP~\cite{hendrycks2016baseline} and MLS~\cite{hendrycks2022scaling} on a well-trained closed-set classifier could achieve competitively with or even outperform state-of-the-art OSR methods. Cen \etal~\cite{cen2023the} further revealed that wrong-classified ID samples have a similar uncertainty distribution to unknown samples, which limits the performance of OSR, and pre-training could enhance both the separation between ID and OOD samples and the separation between the correctly- and wrongly-classified ID samples.

\textbf{\emph{2) Generative Models.}}
Generative models provide different perspectives from discriminative models and could be utilized mainly in two ways, \ie, one is generative and probabilistic modeling of known and unknown classes, and the other is to explicitly generate fake unknown samples for greater separation between known and unknown classes. From the perspective of generative modeling, the basic idea is to model posterior distributions in the latent space to approximate conditional Gaussian distributions~\cite{sun2020conditional}, and the probability of test samples could be used for unknown rejection. However, directly modeling the distribution is difficult, and most existing methods employ off-the-shelf generative models to generate pseudo-unknown samples to augment classifiers. For example, G-OpenMax~\cite{ge2017generative} generates unknown samples with GANs~\cite{goodfellow2014generative} and fine-tunes the model with OpenMax~\cite{bendale2016towards} loss. 
OpenGAN~\cite{kong2021opengan} adopts an adversarial training framework to generate pseudo-unknown samples on the deep feature space.
By further taking the difficulty~\cite{moon2022difficulty} of the generated unknown samples into consideration, models could grasp fine-grained concepts of unknowns for results. Besides, without using explicit generative models, pseudo-unknown samples could be generated by optimizing a latent vector to have a small distance to training samples but produce low confidence on known categories~\cite{neal2018open}. These samples could be viewed as difficult open-space samples that are similar to known samples. A $(K+1)$-way classifier is trained on the combination of training samples and the generated counterfactual samples.

\textbf{\emph{3) Hybrid Models.}}
Discriminative and generative models are complementary to each other, and both have their specific advantages. In principle, discriminative models perform better on classification tasks than generative models. On the other hand, generative models could formalize the distribution of known and unknown samples, and additionally generate fake samples to augment discriminative models. As a consequence, combining both discriminative and generative models into hybrid models results in enormous potential for performing OSR tasks. Zhang \etal~\cite{zhang2020hybrid} proposed to train a flow-based generative model~\cite{chen2019residual} and an inlier classifier on a shared feature space of the encoder. The flow-based model and the classifier are targeted to detect outliers and classify inliers, respectively, without interfering with each other. The flow-based model is trained on the feature space using the maximum likelihood loss function, which avoids the issue of assigning OOD with high probability~\cite{kirichenko2020normalizing} when learned in the input pixel space. Yang \etal~\cite{yang2018robust,yang2020convolutional} creatively proposed convolutional prototype networks (CPN) for OSR with learnable class-wise prototypes serving as a classifier, which is different from conventional linear classifiers. The prototypical classifier could be viewed as Gaussian generative modeling for each class, and leaves more room for open set samples in the feature space. CPN is trained on both discriminative and generative loss functions to encourage inter-class separation and intra-class compactness, respectively, resulting in less open set risk and remarkable OSR performance. ARPL~\cite{chen2020learning,chen2021adversarial} takes a similar spirit of CPN~\cite{yang2020convolutional}, but models class-wise reciprocal points representing the extra-class space, \ie, modeling the open space of each category. ARPL~\cite{chen2021adversarial} additionally employs adversarial enhancement to generate diverse confusing samples to further reduce the open space risk caused by confusing samples. The class-wise points of CPN~\cite{yang2020convolutional} and ARPL~\cite{chen2021adversarial} could be generalized to class-wise continuous manifolds~\cite{huang2022class}, which are constructed by class-wise auto-encoders upon the feature space. Each class-wise manifold could be viewed as an infinite number of prototype/reciprocal points, and thus more flexible and generalizable.

\subsection{Theoretical Study of Unknown Rejection}
\label{subsec:reject-analysis}
Despite numerous empirical approaches for unknown rejection, theoretical investigations remain limited.
Regarding the difficulty of unknown rejection, existing works predominantly resort to \emph{openness}~\cite{scheirer2012toward} $O^\star$ defined by the number of known and unknown classes:
\begin{equation}
    O^\star=1-\sqrt{\frac{2\times K_\text{ID}}{K_\text{ID}+K_\text{OOD}}},
\end{equation}
where $K_\text{ID},K_\text{OOD}$ denote the number of known and unknown classes, respectively. The more unknown categories compared to the number of known categories, the more challenging the unknown rejection task becomes. Vaze \etal~\cite{vaze2022openset} proposed a more rational metric, namely semantic similarity~\cite{vaze2022the}. Open-set samples that share a greater number of visual features with known training samples, such as attributes, color, and shape, are more difficult to reject and detect. 
Fang \etal~\cite{fang2022out} regarded OOD detection as a $(K+1)$-class classification task where $K$ is the number of known class, and then established a theoretical framework for the probably approximately correct learnability of OOD detection. This study finds a necessary condition for the learnability of OOD detection, and presents the impossibility theorems when in-distribution and out-of-distribution overlap with each other. Cheng \etal~\cite{cheng2024breaking} explored whether and how expanding the training set using generated data can improve unknown rejection performance and offered several new insights. The authors established theoretical results that increasing ID generated training data can largely benefit OOD detection under the case where OOD inputs are roughly equally distant from all classes, while having little impact for case where OOD inputs are substantially closer to one class relative to the others. Zhu \etal~\cite{zhu2022rethinking, zhu2023revisiting} revealed an
interesting reliable overfitting phenomenon, \ie, the unknown rejection performance can be easily overfitting during the training of a model. Then, Cheng \etal~\cite{cheng2025average} theoretically show that LASSO \cite{tibshirani1996regression} largely eliminates the
reliable overfitting based on a linear model.

There are also some works focusing on the relationship between OOD detection and misclassification detection. Specifically, it has been observed that many unknown rejection methods excel at detecting OOD samples but sacrifice the ability to filter out misclassified in-distribution examples \cite{zhu2023openmix, jaeger2022call, zhu2022rethinking}, leaving models vulnerable in practical scenarios. To understand this phenomenon, Zhu \etal~\cite{zhu2022rethinking} demonstrated the misalignment between the rejection rules of misclassified ID samples and OOD samples by investigating the Bayes-optimal reject rules. Narasimhan \etal~\cite{narasimhanplugin, narasimhan2023learning} provided a theoretical analysis of the trade-off between OOD detection and misclassification detection.

\subsection{Evaluation of Unknown Rejection}
\label{subsec:evaluation-ur}
The most widely used metric for OOD detection is the Area Under the Receiver Operating Characteristic curve~\cite{davis2006relationship} (AUROC), which is a threshold-independent metric that can be interpreted as the probability that a positive example (ID) is assigned a higher prediction score than a negative example. 
In addition, there are many other metrics like FPR95, AUPR~\cite{hendrycks2016baseline} to imply how the ID and OOD samples are separated. Specifically, AUPR is the Area under the precision–recall (PR) curve. The PR curve is a graph showing the precision versus recall. AUPR typically regards the ID samples as positive. FPR95 can be interpreted as the probability that a negative example is misclassified as a correct prediction when the true positive rate (TPR) is as high as $95\%$. The true positive rate can be computed by TPR=TP/(TP+FN), where TP and FN denote the number of true positives and false negatives, respectively. 
We provide a comprehensive performance comparison of OOD detection and OSR in Table~\ref{tab:ood} and Table~\ref{tab:osr}. Specifically, for OOD detection, the ID datasets are CIFAR-10 and CIFAR-100, and the results are the average on the six common benchmarks \cite{cheng2024breaking, zhu2022rethinking}: Textures \cite{cimpoi2014describing}, SVHN \cite{Netzer2011ReadingDI}, Place365 \cite{ZhouLKO018}, LSUN-C, LSUN-R \cite{YuZSSX15} and iSUN \cite{XuEZFKX15}. From the results in Table~\ref{tab:ood}, we can observe that outlier-aided methods like OE \cite{hendrycks2018deep} perform the best, which is reasonable since OE uses extra natural outlier images. Training-stage methods such as LogitNorm \cite{wei2022mitigating} and GenData \cite{cheng2024breaking} outperform various post-hoc approaches. 
For OSR, the evaluations are conducted on five datasets, including MNIST~\cite{lecun1998gradient}, SVHN~\cite{netzer2011reading}, CIFAR~\cite{krizhevsky2009learning} and TinyImageNet~\cite{le2015tiny}. Recent hybrid methods like ARPL~\cite{chen2021adversarial} and CSSR~\cite{huang2022class} achieve relatively remarkable performance, considering that they combine the advantages of both generative and discriminative models.

\begin{table}[t]
\centering
\setlength\tabcolsep{2pt}
\caption{Comparative results of several OOD methods on CIFAR-10/100, and the results are the average on the six common benchmarks. Here, \textbf{bold} and \underline{underline} indicate the best and the second-best results.}
\vskip -0.05in
\label{tab:ood}
\resizebox{\linewidth}{!}{
\begin{tabular}{ll ccc ccc}
\toprule
\multirow{2}{*}{Method} & \multirow{2}{*}{Venue} & \multicolumn{3}{c}{ID dataset: CIFAR-10} & \multicolumn{3}{c}{ID dataset: CIFAR-100} \\
\cmidrule(l){3-5} \cmidrule(l){6-8}
& & AUROC↑ & AUPR↑ & FPR95↓ & AUROC↑ & AUPR↑ & FPR95↓ \\
\midrule
& & \multicolumn{6}{c}{\textit{Model: ResNet-18}} \\
\cmidrule(l){2-8}
MSP \cite{hendrycks2016baseline} & ICLR 2017 & 92.40 & 98.21 & 47.51 & 80.88 & 95.45 & 74.46 \\
Energy \cite{liu2020energy} & NeurIPS 2020 & 94.32 & 98.49 & 28.78 & 78.21 & 94.62 & 79.89 \\
MaxLogit \cite{hendrycks2022scaling} & ICML 2022 & 94.25 & 98.52 & 29.89 & 78.03 & 94.58 & 79.97 \\
ViM \cite{wang2022vim} & CVPR 2022 & 94.41 & 98.75 & 29.62 & 80.57 & 95.44 & 74.52 \\
KNN \cite{sun2022out} & ICML 2022 & 94.09 & 98.77 & 38.52 & 78.47 & 94.68 & 74.98 \\
LogitNorm \cite{wei2022mitigating} & ICML 2022 & 94.63 & \underline{98.87} & 31.44 & 81.70 & 95.37 & \underline{64.79} \\
RegMixup \cite{pinto2022using} & NeurIPS 2022 & 93.03 & 98.42 & 39.84 & 78.22 & 94.78 & 81.73 \\
OE \cite{hendrycks2018deep} & ICLR 2019 & \textbf{98.61} & \textbf{99.71} & \textbf{4.78} & \textbf{89.70} & \textbf{97.72} & \textbf{52.98} \\
MSP++ \cite{vaze2022openset} & ICLR 2022 & 93.21 & 98.36 & 41.36 & 77.03 & 94.36 & 80.38 \\
GenData \cite{cheng2024breaking} & IJCV 2024 & \underline{96.27} & \underline{99.21} & \underline{21.32} & \underline{84.24} & \underline{96.39} & 66.07 \\
\midrule
& & \multicolumn{6}{c}{\textit{Model: WRN-28-10}} \\
\cmidrule(l){2-8}
MSP \cite{hendrycks2016baseline} & ICLR 2017 & 91.11 & 97.34 & 41.76 & 77.80 & 94.40 & 79.48 \\
Energy \cite{liu2020energy} & NeurIPS 2020 & 92.46 & 97.48 & 29.19 & 76.87 & 93.91 & 79.54 \\
MaxLogit \cite{hendrycks2022scaling} & ICML 2022 & 92.43 & 97.50 & 30.02 & 76.67 & 93.90 & 79.71 \\
ViM \cite{wang2022vim} & CVPR 2022 & 96.34 & 99.11 & 17.07 & 84.36 & 96.14 & \underline{62.42} \\
KNN \cite{sun2022out} & ICML 2022 & 94.73 & 98.91 & 33.38 & 79.34 & 94.95 & 75.89 \\
LogitNorm \cite{wei2022mitigating} & ICML 2022 & \underline{97.36} & \underline{99.42} & \underline{13.26} & \underline{85.27} & \underline{96.50} & 64.61 \\
RegMixup \cite{pinto2022using} & NeurIPS 2022 & 93.16 & 98.29 & 36.92 & 83.13 & 95.93 & 69.38 \\
OE \cite{hendrycks2018deep} & ICLR 2019 & \textbf{98.65} & \textbf{99.74} & \textbf{3.69} & \textbf{91.43} & \textbf{98.18} & \textbf{49.11} \\
MSP++ \cite{vaze2022openset} & ICLR 2022 & 91.83 & 97.45 & 36.16 & 76.42 & 93.90 & 80.00 \\
GenData \cite{cheng2024breaking} & IJCV2024  &96.83 & 99.21 &15.18 & 84.93 & 96.45 & 64.76 \\
\bottomrule
\end{tabular}
}
\end{table}

\begin{table}[!t]
\setlength\tabcolsep{2pt}
\centering
\renewcommand{\arraystretch}{1}
\caption{Comparative results of several OSR methods with AUROC.}
\vskip -0.05in
\label{tab:osr}
\resizebox{\linewidth}{!}{
\begin{tabular}{@{}llcccccc@{}}
\toprule
Method & Venue & MNIST & SVHN & CIFAR10 & CIFAR+10 & CIFAR+50 & TinyImageNet \\ \midrule
MSP~\cite{hendrycks2016baseline} & Baseline & 97.8 & 88.6 & 67.7 & 81.6 & 80.5 & 57.7 \\
OpenMax~\cite{bendale2016towards} & CVPR 2016 & 98.1 & 89.4 & 69.5 & 81.7 & 79.6 & 57.6 \\
G-OpenMax~\cite{ge2017generative} & BMVC 2017 & 98.4 & 89.6 & 67.5 & 82.7 & 81.9 & 58.0 \\
OSRCI~\cite{neal2018open} & ECCV 2018 & 98.8 & 91.0 & 69.9 & 83.8 & 82.7 & 58.6 \\
CSOSR~\cite{yoshihashi2019classification} & CVPR 2019 & 99.1 & 89.9 & 88.3 & 91.2 & 90.5 & 58.9 \\
C2AE~\cite{oza2019c2ae} & CVPR 2019 & 98.9 & 92.2 & 89.5 & 95.5 & 93.7 & 74.8 \\
PROSER~\cite{zhou2021learning} & CVPR 2021 & 96.4 & 94.3 & 89.1 & 96.0 & 95.3 & 69.3 \\
OpenHybrid~\cite{zhang2020hybrid} & ECCV 2020 & 99.5 & 94.7 & 95.0 & 96.2 & \underline{95.5} & 79.3 \\
DIAS~\cite{moon2022difficulty} & ECCV 2022 & 99.2 & 94.3 & 85.0 & 92.0 & 91.6 & 73.1 \\
CPN~\cite{yang2020convolutional} & TPAMI 2020 & 99.0 & 92.6 & 82.8 & 88.1 & 87.9 & 63.9 \\
ARPL~\cite{chen2021adversarial} & TPAMI 2021 & \underline{99.6} & 96.3 & 90.1 & \underline{96.5} & 94.3 & 76.2 \\
ARPL+CS~\cite{chen2021adversarial} & TPAMI 2021 & \textbf{99.7} & \underline{96.7} & \underline{91.0} & \textbf{97.1} & 95.1 & \underline{78.2} \\
CSSR~\cite{huang2022class} & TPAMI 2022 & - & \textbf{97.9} & \textbf{91.3} & 96.3 & \textbf{96.2} & \textbf{82.3} \\ \bottomrule
\end{tabular}
}
\end{table}


\section{Novel Class Discovery}
\label{sec:novel-discovery}

\subsection{Problem Formulation and Key Challenges}
\textbf{\emph{1) Problem Formulation.}} Novel class discovery~\cite{han2021autonovel,troisemaine2023novel} is the second step in open-world machine learning, which aims to automatically discover novel categories from unlabeled data based on the model's prior knowledge. Novel class discovery is an extension of unknown rejection, requiring models to not only reject unknown samples, but also further extrapolate to classify the rejected samples.
Given a dataset $\mathcal{D}_\text{train}=\mathcal{D}_l\cup \mathcal{D}_u$ with two subsets, where $\mathcal{D}_l=\{(\mathbf{x}_i,y_i)\}_{i=1}^{N_l}\subset \mathcal{X}\times \mathcal{Y}_l$ is the labeled dataset and $\mathcal{D}_u=\{(\mathbf{x}_i,y_i)\}_{i=1}^{N_u}\subset \mathcal{X}\times \mathcal{Y}_u$ is the unlabeled data. It is worth noting that $\mathcal{Y}_u\neq \mathcal{Y}_l$, \ie, there are novel classes in the unlabeled data outside the old classes in $\mathcal{D}_l$. Let $\mathcal{C}_{\text{old}},\mathcal{C}_{\text{new}}$ denote the old and new classes in the labeled and unlabeled dataset, with the number of classes $K_{\text{old}}=|\mathcal{C}_{\text{old}}|, K_{\text{new}}=|\mathcal{C}_{\text{new}}|$, respectively. The objective is to cluster and discover $\mathcal{C}_{\text{new}}$ in $\mathcal{D}_u$ leveraging the knowledge learned on $\mathcal{D}_l$.

In essence, novel class discovery is a clustering task on $\mathcal{D}_u$. Regarding its differences from unsupervised clustering~\cite{xie2016unsupervised,caron2018deep}, the latter aims to cluster unlabeled data in a purely unsupervised manner without any prior knowledge of the classification criterion. As a result, unsupervised clustering is not a fully learnable task. For example, in the absence of any external information, the model might face a dilemma, for example, being confused about whether to cluster red flowers and red cars together according to colors or to cluster red flowers and yellow flowers together according to species. From this perspective, some labeled data $\mathcal{D}_l$ is necessary and could provide some prior knowledge indicating what constitutes a class to clarify the clustering criterion and eliminate ambiguity. In a word, novel class discovery is a more reasonable and pragmatic task, and could be referred to as deep transfer clustering~\cite{han2019learning}, in which models are supposed to transfer the knowledge learned from old and labeled classes to cluster novel categories in the unlabeled data.

\textbf{\emph{2) Key Challenges.}} Intrinsically, the semantic similarity between new and old categories determines the extent of knowledge transferable from old to new categories, thereby influencing the performance of category discovery. If the semantic similarity is limited, the labeled old class might degrade the performance of novel class discovery~\cite{li2023supervised}. Technically, novel category discovery could be referred to as open-world semi-supervised learning~\cite{cao2022openworld}, where unlabeled samples contain new classes, which makes the conventional pseudo-labeling mechanism~\cite{lee2013pseudo} not applicable due to the non-shared categories between labeled and unlabeled data. Moreover, in practical scenarios, the number of novel categories is always unknown, which makes the task challenging.

This section will sequentially introduce two primary tasks of novel class discovery, namely novel category discovery (Sec.~\ref{subsec:ncd}) and generalized category discovery (Sec.~\ref{subsec:gcd}), along with representative methods. We also elaborate on the class number estimation issue (Sec.~\ref{subsec:number-estimation}), which is critical in realistic scenarios. Besides, we point out some extended settings of novel class discovery to reflect some new trends of this field (Sec.~\ref{subsec:ncd-extend}). Last, we summarize theoretical results of category discovery (Sec.~\ref{subsec:ncd-analysis}) and then present the evaluation protocols and performance comparisons (Sec.~\ref{subsec:eval}).

\begin{figure*}[t]
  \begin{center}
\centerline{\includegraphics[width=\textwidth]{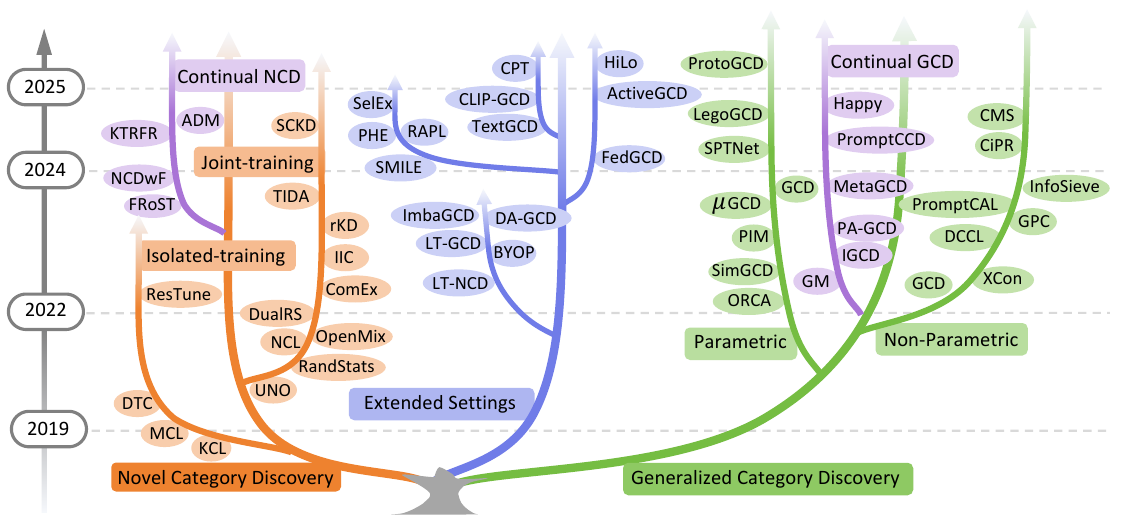}}
   \caption{The evolutionary tree of novel class discovery methods.}
   \label{fig:tree-ncd}
 \end{center}
 \vskip -0.2in
\end{figure*}

\subsection{Novel Category Discovery}
\label{subsec:ncd}

Novel Category Discovery~\cite{han2021autonovel} (NCD) is a basic setting where $\mathcal{Y}_u=\mathcal{C}_{\text{new}}$ and $\mathcal{Y}_u\bigcap\mathcal{Y}_l=\Phi$, \ie, unlabeled data only contain novel classes and has no class overlap with the labeled data. NCD mainly concerns category discovery in $\mathcal{D}_u$.

\textbf{\emph{1) Isolated-training Methods.}}
Early works in NCD mainly transfer knowledge from $\mathcal{D}_l$ to $\mathcal{D}_u$ for category discovery in isolated steps. They first focus on $\mathcal{D}_l$ only to learn the similarity function. Subsequently, they utilize the learned similarity function to perform unlabeled learning exclusively on the unlabeled data $\mathcal{D}_u$. In other words, these methods do not learn new and old classes simultaneously. Instead, they learn them at separate stages. A pioneering work, DTC~\cite{han2019learning}, formalizes NCD as a deep transfer clustering task, and builds the method upon unsupervised clustering DEC~\cite{xie2016unsupervised}. DTC first trains a model on $D_l$ in a supervised manner, which is then utilized to extract features of $\mathcal{D}_u$ to initialize cluster centroids. Then, DTC performs alternative optimization between the soft targets and model parameters, as well as the cluster centroids. Later works predominantly resort to pseudo-labels to transfer knowledge from labeled samples from old classes to unlabeled samples from novel classes. However, due to the disjoint classes, one could not simply adopt the naive pseudo-labels~\cite{lee2013pseudo} in semi-supervised learning~\cite{NEURIPS2018_c1fea270}. To overcome this issue, several methods rely on pairwise pseudo-labels~\cite{hsu2018learning,hsu2018multiclass} indicating whether two samples are similar or not. Specifically, the model is first trained on $\mathcal{D}_l$ to assign pairwise pseudo-labels on $\mathcal{D}_u$, if the similarity is larger than a pre-defined threshold, then such pairs of samples are recognized as similar pairs. Then, the model is trained on $\mathcal{D}_u$ with the binary pseudo-labels for novel category discovery. Such types of methods rely on an implicit assumption that general knowledge of what constitutes a good class is shared across old and new classes. They assume that the similarity function learned from $\mathcal{D}_l$ is also applicable to $\mathcal{D}_u$. ResTune~\cite{liu2022residual} disentangles the representations of old and new classes, which facilitates the adaptation of new classes and mitigates the forgetting of old classes.
\begin{figure*}[!t]
	\centering
	\includegraphics[width=\textwidth]{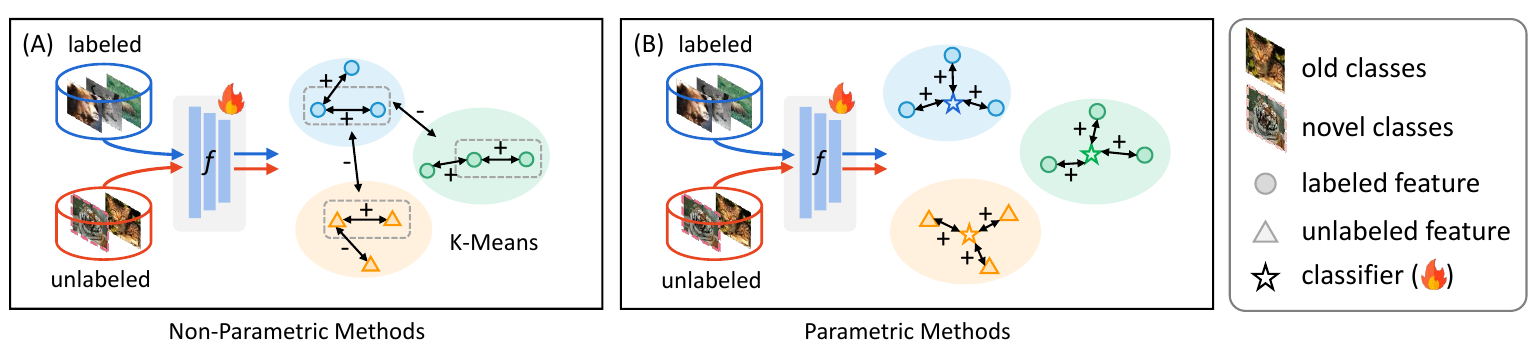}
	\vskip -0.1in
	\caption{Illustrations of two types of methods of Generalized Category Discovery (GCD), \ie, (A) non-parametric methods with K-means clustering and (B) parametric methods with learnable classification heads.}
	\label{fig:gcd}
\end{figure*}

\textbf{\emph{2) Joint-training Methods.}}
The fundamental principle of these methods lies in training the model simultaneously on both labeled $\mathcal{D}_l$ and unlabeled data $\mathcal{D}_u$. Although some of them~\cite{han2021autonovel,zhao2021novel} employ multi-stage training, they utilize labeled data of old classes $\mathcal{D}_l$ during the clustering of unlabeled new classes in $\mathcal{D}_u$, rather than treating the learning of new and old classes as isolated, independent tasks, as in the isolated-training methods.
AutoNovel~\cite{han2021autonovel} is a representative work in NCD, dividing NCD into three steps. The model is first pre-trained on $\mathcal{D}_l$ and $\mathcal{D}_u$ collectively with self-supervision on rotation prediction~\cite {gidaris2018unsupervised} to grasp fundamental representation. They proposed a novel pseudo-labeling criteria, namely ranking statistics to obtain pairwise similarities $s_{ij}$. Concretely, two samples are deemed similar if the top-k ranked dimensions of feature magnitudes are identical, which is robust and stable for knowledge transfer from old to novel classes, as in Eq.~\eqref{eq:rank-stat}.
\begin{equation}
    s_{ij}=\mathds{1}(\text{top}_k \boldsymbol{z}_i^u=\text{top}_k \boldsymbol{z}_j^u),
    \label{eq:rank-stat}
\end{equation}
where $\boldsymbol{z}_i^u$ denotes the visual feature of the $i$-th unlabeled sample. In the last step, the model is jointly optimized on conventional cross-entropy loss on $\mathcal{D}_l$ and binary cross-entropy loss (Eq.~\eqref{eq:loss-bce}) on $\mathcal{D}_u$ with pair-wise pseudo-labels, together with consistency regularization~\cite{sajjadi2016regularization} (Eq.~\eqref{eq:loss-mse}) for better representation learning.
\begin{equation}
\begin{aligned}
    \mathcal{L}_\text{bce}=-\frac{1}{|\mathcal{B}|^2}\sum_{i\in\mathcal{B}}\sum_{j\in\mathcal{B}} \big[s_{ij}\log \eta^u(\boldsymbol{z}_i^u)^\top\eta^u(\boldsymbol{z}_j^u) \\
    +(1-s_{ij})\log(1-\eta^u(\boldsymbol{z}_i^u)^\top\eta^u(\boldsymbol{z}_j^u))\big],
    \label{eq:loss-bce}
\end{aligned}
\end{equation}
\begin{equation}
\begin{aligned}
    \mathcal{L}_\text{mse} & = \frac{1}{|\mathcal{B}_l|}\sum_{i\in\mathcal{B}_l}\Vert \eta^l(\boldsymbol{z}_i^l)-\eta^l(\hat{\boldsymbol{z}}_i^l)\Vert^2\\
    & + \frac{1}{|\mathcal{B}_u|}\sum_{i\in\mathcal{B}_u}\Vert \eta^l(\boldsymbol{z}_i^u)-\eta^l(\hat{\boldsymbol{z}}_i^u)\Vert^2,
    \label{eq:loss-mse}
\end{aligned}
\end{equation}
where $\eta^l(\cdot),\eta^u(\cdot)$ denotes the classification head for old and new classes, respectively, and $\boldsymbol{z}_i$ is the feature representation, while $\hat{\boldsymbol{z}}_i$ is another view with a different data augmentation. Some other methods advance NCD from different perspectives, including local part-level information~\cite{zhao2021novel}, and neighborhood~\cite{Zhong_2021_CVPR} in the feature space containing more potential positive samples. The adjacent samples in the feature space are valuable for constructing pseudo-positives and hard negatives for substantially better representation learning.
Zhong \etal~\cite{zhong2021openmix} model old and novel classes in a joint label distribution, and introduce OpenMix to MixUp~\cite{zhang2018mixup} the labeled and unlabeled samples, as well as generating pseudo-labels by mixing respective pseudo and ground-truth labels, which are more credible for learning. UNO~\cite{Fini_2021_ICCV} is a simple yet effective method with a unified objective for NCD. UNO employs a swapped prediction task between two views motivated by SwAV~\cite{caron2020unsupervised}. The pseudo targets are derived through the Sinkhorn-Knopp algorithm~\cite{cuturi2013sinkhorn}, which encourages equipartition of pseudo-labels across various classes. The unlabeled self-learning and supervised learning on labeled data are collectively implemented in an end-to-end manner. ComEx~\cite{yang2022divide} divides and conquers NCD with two groups of compositional experts, with each characterizing the whole dataset from complementary perspectives. Li \etal~\cite{Li_2023_CVPR} propose to utilize Kullback–Leibler divergence terms to maximize the inter-class divergence and minimize the intra-class divergence, achieving strong performance.

\subsection{Generalized Category Discovery}
\label{subsec:gcd}

Generalize category discovery (GCD) is a more pragmatic and challenging task, relaxing the strong assumption in NCD that unlabeled data all come from novel classes, \ie, $\mathcal{Y}_u=\mathcal{C}_{\text{old}}\bigcap\mathcal{C}_{\text{new}}$ and $\mathcal{Y}_l\subset \mathcal{Y}_u$. The challenge in GCD lies in that models are supposed to simultaneously recognize unlabeled samples from old classes and cluster samples from novel categories. Illustrations of two canonical types of methods are shown in Fig.~\ref{fig:gcd}.

\textbf{\emph{1) Non-parametric Classification.}}
Prior methods~\cite{vaze2022gcd,fei2022xcon,Zhao_2023_ICCV} in GCD predominantly rely on contrastive learning for feature representation. They do not explicitly learn a parametric classification head. Instead, they resort to K-means~\cite{macqueen1967some,arthur2007k} clustering as a non-parametric classifier for classification. Vaze \etal~\cite{vaze2022gcd} firstly introduced the task of GCD, and proposed to employ supervised contrastive learning~\cite{khosla2020supervised} on labeled data $\mathcal{D}_l$ (Eq.~\eqref{eq:loss-sup-con}) and unsupervised contrastive learning~\cite{chen2020simple} on all data $\mathcal{D}$ (Eq.~\eqref{eq:loss-unsup-con}), respectively. 
\begin{equation}
    \mathcal{L}_\text{con}^l=\frac{1}{|\mathcal{B}_l|}\sum_{i\in\mathcal{B}_l}\frac{1}{|\mathcal{N}(i)|}\sum_{q\in\mathcal{N}(i)}-\log\frac{\exp(\mathbf{h}_i^\top \mathbf{h}_q/\tau_c)}{\sum_{j}\mathds{1}_{[j\neq i]}\exp(\mathbf{h}_i^\top \mathbf{h}_j/\tau_c)},
    \label{eq:loss-sup-con}
\end{equation}
\begin{equation}
    \mathcal{L}_\text{con}^u=\frac{1}{|\mathcal{B}|}\sum_{i\in\mathcal{B}}-\log \frac{\exp(\mathbf{h}_i^\top \mathbf{h}_i^\prime/\tau_c)}{\sum_j \mathds{1}_{[j\neq i]}\exp(\mathbf{h}_i^\top \mathbf{h}_j/\tau_c)}.
    \label{eq:loss-unsup-con}
\end{equation}
Here, $\mathcal{B}_l$ and $\mathcal{B}$ denote labeled samples and all samples in a batch, respectively. $\mathbf{h}$ represents the feature in projection space for contrastive learning. Then, semi-supervised K-means~\cite{macqueen1967some} is performed on the learned representations for the classification of old samples and clustering of novel samples. They argue that parametric classifiers are prone to overfitting to old classes. This work could be further augmented by performing contrastive learning on sub-datasets partitioned by clustering in advance~\cite{fei2022xcon}. However, purely unsupervised contrastive learning neglects substantial positive pairs and suffers from the \emph{class collision} issue. Later works exploit more potential positive pairs to better feature representation. For example, a dynamic conceptual contrastive learning~\cite{pu2023dynamic} framework is introduced to exploit the underlying relationships between samples of similar concepts. In this framework, dynamic conception generation and conception-level contrastive learning are alternatively performed for better feature representations. Zhao \etal~\cite{Zhao_2023_ICCV} integrated class number estimation and representation learning into a unified EM-like framework to learn semi-supervised Gaussian mixture models. To further achieve better performance, PromptCAL~\cite{zhang2023promptcal} employs visual prompt tuning~\cite{jia2022visual} with stronger representation ability from a diagonal perspective. InfoSieve~\cite{rastegar2023learn} conceptualizes a category through the view of optimization and reformulates the task of category discovery as the assignment of minimum-length category codes for each sample. Hao \etal~\cite{hao2024cipr} take the spirit of agglomerative clustering~\cite{murtagh2014ward} and construct a clustering hierarchy from the graph of selective neighbors.

\textbf{\emph{2) Parametric Classification.}}
Contrary to~\cite{vaze2022gcd}, Wen \etal~\cite{Wen_2023_ICCV} rethink parametric classifiers for GCD and attribute the failure of parametric classifiers to two biases, \ie, the bias to predicting old classes more often and the bias to imbalanced pseudo-labels. They utilize self-distillation~\cite{caron2021emerging} between two augmentation views for self-training, as shown below:
\begin{equation}
    \mathcal{L}_\text{distill}=\frac{1}{|\mathcal{B}|}\sum_{i\in\mathcal{B}}\ell(\boldsymbol{q}_i^\prime,\boldsymbol{p}_i).
\end{equation}
Here $\ell(\cdot)$ is the cross-entropy loss and $\boldsymbol{q}_i^\prime$ is the soft target from the other view with a lower temperature, \ie, more confident distributions. Furthermore, entropy regularization is employed to mitigate imbalanced pseudo-labels and biased predictions toward old classes:
\begin{equation}
    \mathcal{L}_\text{entropy}=-H(\overline{\boldsymbol{p}})=\sum_{k=1}^K \overline{\boldsymbol{p}}^{(k)}\log \overline{\boldsymbol{p}}^{(k)},
    \label{eq:loss-entropy}
\end{equation}
where $K$ is the total number of classes, and $H(\boldsymbol{p})=-\sum_{k}\boldsymbol{p}^{(k)}\log \boldsymbol{p^{(k)}}$ denotes entropy, and $\overline{\boldsymbol{p}}=\frac{1}{|\mathcal{B}|}\sum_{i\in\mathcal{B}}\boldsymbol{p}_i$ denotes marginal probability. By avoiding severe biases, the proposed method outperforms non-parametric methods by a large margin. $\mu$GCD~\cite{vaze2023no} incorporates more advanced training schemes in FixMatch~\cite{sohn2020fixmatch}, \eg, exponential moving average and misaligned data augmentations to obtain more remarkable performance. SPTNet~\cite{wang2024sptnet} proposes spatial prompt learning shared across images, empowering the model to focus on object parts. ProtoGCD~\cite{10948388} proposes a hybrid modeling with both discriminative and generative learning. Specifically, ProtoGCD models class-wise probability distribution with learnable prototypes. To balance the noise and self-learning efficiency of pseudo-labeling, dual-level adaptive pseudo-labels are introduced to consider the adaptivity across different training samples and phases. ProtoGCD also explicitly regularizes the class-wise prototypes to be more separable. Consequently, ProtoGCD achieves remarkable representations with large inter-class separation and intra-class compactness. The authors~\cite{10948388} also provide theoretical support for the design of pseudo-labels. Besides, PIM~\cite{chiaroni2023parametric} provides a theoretical analysis for parametric methods from the perspective of mutual information. Generally, parametric classification methods are more efficient with end-to-end training. For inference, models could directly produce predictions with parametric classification heads without further clustering.

\subsection{Estimating the Number of Novel Classes}
\label{subsec:number-estimation}

In the literature of NCD and GCD, most methods require the number of novel classes $|\mathcal{C}_{\text{new}}|$ to be known \emph{a priori}, which is impractical in real-world applications. It is of vital importance to estimate the number of classes precisely. AutoNovel~\cite{han2021autonovel} devises a sophisticated data splitting and clustering scheme. In particular, they first split the probe (labeled) data $\mathcal{D}_l^r$ into $\mathcal{D}_l^{ra}$ and $\mathcal{D}_l^{rv}$, and run semi-supervised K-means on $\mathcal{D}_l^r\bigcup\mathcal{D}_u$ with ground-truth constraints on $\mathcal{D}_l^{ra}$, subsequently, they proposed to compute clustering accuracy on $\mathcal{D}_l^{rv}$ and cluster validity index (CVI)~\cite{arbelaitz2013extensive} on $\mathcal{D}_u$, respectively. Finally, they average the cluster number corresponding to the maximum accuracy and CVI value, as the estimation of the class number. Vaze \etal~\cite{vaze2022gcd} proposed to run the clustering algorithm on the learned representations with different numbers of classes and choose the one that maximizes the accuracy on the labeled data as an estimate of $|\mathcal{C}_{\text{new}}|$. Recent methods adopt a more elegant way, which seamlessly integrates representation learning and category number estimation into a unified learning framework. DCCL~\cite{pu2023dynamic} utilizes the infomap algorithm~\cite{rosvall2009map} for dynamic conception generation, where the number of conceptions is dynamically changed throughout training. As the algorithm converges, the resulting number of clusters represents the estimated results. Zhao \etal~\cite{Zhao_2023_ICCV} formalized an EM-like framework, wherein prototype and class number are estimated in the E-step with a variant of the Gaussian mixture model, and they employ prototypical contrastive learning~\cite{li2021prototypical} in the M-step. A more recent work, ProtoGCD~\cite{10948388}, devises \emph{prototype score}, which simultaneously considers the accuracy of labeled data and the characteristics of feature space and achieves more precise estimation results.

\subsection{Extended Settings}
\label{subsec:ncd-extend}

Recent efforts have extended GCD to other machine learning paradigms, like active learning~\cite{ma2024active} and federated learning~\cite{pu2024federated}. Ma \etal~\cite{ma2024active} uncover the inherent imbalance problem between old and new classes in GCD. To effectively overcome this issue, they introduce the task of active generalized category discovery. By actively selecting valuable unlabeled samples, potentially from new classes and querying their labels from the oracle, the limited accuracy of new classes could be remarkably enhanced. To meet the privacy and decentralization requirements in practical scenarios, Pu \etal~\cite{pu2024federated} formalize federated generalized category discovery. Furthermore, by considering the realistic streaming data, GCD could also be extended to the continual learning setting, \ie, continual category discovery~\cite{joseph2022novel,roy2022class,Kim_2023_ICCV,Zhao_2023_ICCV_Incremental,NEURIPS2024_5ae0f7cf}, which could also be referred to as unsupervised continual learning. We will cover this content in Section~\ref{subsec:cil-extend}. Beyond the single visual modality, recent works~\cite{zheng2024textual,wang2024get,yang2025consistent} leverage orthogonal information from the text modality to further upgrade the performance of category discovery.

\subsection{Theoretical Analysis of Novel Class Discovery}
\label{subsec:ncd-analysis}

Novelty class discovery is implicitly guided by knowledge transferred from old classes to novel categories for clustering and category discovery. How much knowledge could be transferred essentially influences the performance of category discovery. Li \etal~\cite{li2023supervised} questioned the viewpoint that supervised knowledge is always helpful at different levels of semantic relevance. They proposed a metric called transfer flow to quantitatively assess the semantic similarity between labeled and unlabeled categories and give a theoretical analysis of semantic similarity and conclude that supervised knowledge might hurt the performance of NCD in circumstances with low semantic similarity between old and new classes. They further suggest utilizing transfer flow as a reference to indicate how to use the training data. Generally, the higher the semantic similarity between the old and new classes, the more substantial the contribution of the supervised training data to NCD.

As for the clustering mechanism of GCD, Chiaroni \etal~\cite{chiaroni2023parametric} introduce constrained mutual information:
\begin{equation}
    \max_{\boldsymbol{W}} H(Y)-H(Y|Z),\quad \text{s.t.}\ \boldsymbol{y}_i=\boldsymbol{p}_i,\quad \forall z_i\in Z_l.
    \label{eq:infomax}
\end{equation}
The overall objective is to maximize the marginal entropy $H(Y)$ while minimizing the conditional entropy of labels $Y$ conditioned on the underlying features $Z$. The first term corresponds to the equipartition assumption of class distributions. As for the second term, we need to assign the target $\boldsymbol{h}_i$ for each sample. In the context of GCD, there are both labeled data from old classes, as well as unlabeled samples from all classes. For the labeled samples, the annotations $\boldsymbol{y}_i$ are directly employed. For the unlabeled ones, they propose to use the predictive probabilities $\boldsymbol{p}_i$, which is equal to minimizing the predictive entropy of each sample, \ie, encouraging more confident predictions. In this way, the model performs semi-supervised learning and self-training.

Intuitively, the difficulty of NCD is determined by the semantic similarities between old and novel categories, \ie, how much knowledge is helpful and could be transferred for NCD. Besides, the number of novel classes also influences the difficulty. The greater the semantic similarity and the fewer the novel categories, the simpler the NCD task becomes. Interestingly, the impact of semantic similarity on the difficulty of novel class discovery is opposite to the conclusion of the unknown rejection task, in essence, unknown rejection is a discriminative task to distinguish between ID and OOD samples, while NCD is a transfer learning task to transfer the knowledge from ID to guide the clustering of OOD samples.

\begin{table}[!t]
\setlength\tabcolsep{3pt}
\centering
\renewcommand{\arraystretch}{1}
\caption{Comparative results of several NCD methods using task-aware evaluation protocol.}
\vskip -0.05in
\label{tab:ncd-task-aware}
\resizebox{\linewidth}{!}{
\begin{tabular}{@{}llcccc@{}}
\toprule
Method & Venue & CIFAR10 & CIFAR100-20 & CIFAR100-50 & ImageNet \\ \midrule
k-means~\cite{arthur2007k} & baseline & 72.5 & 56.3 & 28.3 & 71.9 \\
KCL~\cite{hsu2018learning} & ICLR 2018  & 72.3 & 42.1 & - & 73.8 \\
MCL~\cite{hsu2018multiclass} & ICLR2018 & 70.9 & 21.5 & - & 74.4 \\
DTC~\cite{han2019learning} & ICCV 2019 & 88.7 & 67.3 & 35.9 & 78.3 \\
RS~\cite{Han2020Automatically} & ICLR 2020 & 90.4 & 73.2 & 39.2 & 82.5 \\
RS+~\cite{han2021autonovel} & TPAMI 2021 & 91.7 & 75.2 & 44.1 & 82.5 \\
OpenMix~\cite{zhong2021openmix} & CVPR 2021 & 95.3 & - & - & 85.7 \\
NCL~\cite{Zhong_2021_CVPR} & CVPR 2021 & 93.4 & {\ul 86.6} & - & {\ul 90.7} \\
UNO~\cite{Fini_2021_ICCV} & ICCV 2021 & {\ul 96.1} & 85.0 & 52.9 & 90.6 \\
DualRank~\cite{zhao2021novel} & NeurIPS 2021 & 91.6 & 75.3 & - & 88.9 \\
ComEx~\cite{yang2022divide} & CVPR 2022 & 93.6 & 85.7 & {\ul 53.4} & 90.9 \\
IIC~\cite{Li_2023_CVPR} & CVPR 2023 & \textbf{99.1} & \textbf{92.4} & \textbf{65.8} & \textbf{91.9} \\ \bottomrule
\end{tabular}
}
\end{table}

\begin{table}[!t]
\setlength\tabcolsep{1.5pt}
\centering
\renewcommand{\arraystretch}{1.15}
\caption{Comparative results of several NCD methods using task-agnostic evaluation protocol.}
\vskip -0.05in
\label{tab:ncd-task-agnostic}
\resizebox{\linewidth}{!}{
\begin{tabular}{@{}llccccccccc@{}}
\toprule
\multirow{2}{*}{Method} & \multirow{2}{*}{Venue} & \multicolumn{3}{c}{CIFAR10} & \multicolumn{3}{c}{CIFAR100-20} & \multicolumn{3}{c}{CIFAR100-50} \\ \cmidrule(l){3-11} 
 & & Labeled & Unlabeled & All & Labeled & Unlabeled & All & Labeled & Unlabeled & All \\ \midrule
KCL~\cite{hsu2018learning} & ICLR 2018 & 79.4 & 60.4 & 69.8 & 23.4 & 29.4 & 24.6 & - & - & - \\
MCL~\cite{hsu2018multiclass} & ICLR 2018 & 81.4 & 64.8 & 73.1 & 18.2 & 18.0 & 18.2 & - & - & - \\
DTC~\cite{han2019learning} & ICCV 2019 & 58.7 & 78.6 & 68.7 & 47.6 & 49.1 & 47.9 & 30.2 & 34.7 & 32.5 \\
RS+~\cite{han2021autonovel} & TPAMI 2021 & 90.6 & 88.8 & 89.7 & 71.2 & 56.8 & 68.3 & 69.7 & 40.9 & 55.3 \\
UNO~\cite{Fini_2021_ICCV} & ICCV 2021 & 93.5 & {\ul 93.3} & 93.4 & 73.2 & 73.1 & 73.2 & 71.5 & 50.7 & 61.1 \\
ComEx~\cite{yang2022divide} & CVPR 2022 & {\ul 95.0} & 92.6 & {\ul 93.8} & {\ul 75.2} & {\ul 77.3} & {\ul 75.6} & \textbf{75.3} & {\ul 53.5} & {\ul 64.4} \\
IIC~\cite{Li_2023_CVPR} & CVPR 2023 & \textbf{96.0} & \textbf{97.2} & \textbf{96.6} & \textbf{75.9} & \textbf{78.4} & \textbf{77.2} & {\ul 75.1} & \textbf{61.0} & \textbf{68.1} \\ \bottomrule
\end{tabular}
}
\vskip -0.15in
\end{table}

\begin{table*}[!t]
\setlength\tabcolsep{10pt}
\centering
\renewcommand{\arraystretch}{1}
\caption{Comparative results of several GCD methods on generic datasets.}
\label{tab:gcd-generic}
\resizebox{.8\linewidth}{!}{
\begin{tabular}{@{}lllccccccccc@{}}
\toprule
\multirow{2}{*}{Type} & \multirow{2}{*}{Methods} & \multirow{2}{*}{Venue} & \multicolumn{3}{c}{CIFAR10} & \multicolumn{3}{c}{CIFAR100} & \multicolumn{3}{c}{ImageNet-100} \\ \cmidrule(l){4-6} \cmidrule(l){7-9} \cmidrule(l){10-12} 
 &  &  & All & Old & New & All & Old & New & All & Old & New \\ \midrule
\multirow{8}{*}{\begin{tabular}[c]{@{}l@{}}Non-parametric\\ Classification\end{tabular}} & K-means~\cite{arthur2007k} & Baseline & 83.6 & 85.7 & 82.5 & 52.0 & 52.2 & 50.8 & 72.7 & 75.5 & 71.3 \\
 & GCD~\cite{vaze2022gcd} & CVPR 2022 & 91.5 & 97.9 & 88.2 & 73.0 & 76.2 & 66.5 & 74.1 & 89.8 & 66.3 \\
 & XCon~\cite{fei2022xcon} & BMVC 2022 & 96.0 & 97.3 & 95.4 & 74.2 & 81.2 & 60.3 & 77.6 & 93.5 & 69.7 \\
 & DCCL~\cite{pu2023dynamic} & CVPR 2023 & 96.3 & 96.5 & 96.9 & 75.3 & 76.8 & 70.2 & 80.5 & 90.5 & 76.2 \\
 & GPC~\cite{Zhao_2023_ICCV} & ICCV 2023 & 92.2 & {\ul 98.2} & 89.1 & 77.9 & {\ul 85.0} & 63.0 & 76.9 & 94.3 & 71.0 \\
 & PromptCAL~\cite{zhang2023promptcal} & CVPR 2023 & \textbf{97.9} & 96.6 & {\ul 98.5} & 81.2 & 84.2 & 75.3 & 83.1 & 92.7 & 78.3 \\
 & InfoSieve~\cite{rastegar2023learn} & NeurIPS 2023 & 94.8 & 97.7 & 93.4 & 78.3 & 82.2 & 70.5 & 80.5 & 93.8 & 73.8 \\
 & CiPR~\cite{hao2024cipr} & TMLR 2024 & {\ul 97.7} & 97.5 & 97.7 & 81.5 & 82.4 & {\ul 79.7} & 80.5 & 84.9 & 78.3 \\
 & CMS~\cite{choi2024contrastive} & CVPR 2024 & - & - & - & \textbf{82.3} & \textbf{85.7} & 75.5 & {\ul 84.7} & \textbf{95.6} & 79.2 \\ \midrule \midrule
\multirow{6}{*}{\begin{tabular}[c]{@{}l@{}}Parametric\\ Classification\end{tabular}} & RankStats+~\cite{han2021autonovel} & TPAMI 2021 & 46.8 & 19.2 & 60.5 & 58.2 & 77.6 & 19.3 & 37.1 & 61.6 & 24.8 \\
 & UNO+~\cite{Fini_2021_ICCV} & ICCV 2021 & 68.6 & \textbf{98.3} & 53.8 & 69.5 & 80.6 & 47.2 & 70.3 & 95.0 & 57.9 \\
 & ORCA~\cite{cao2022openworld} & ICLR 2022 & 81.8 & 86.2 & 79.6 & 69.0 & 77.4 & 52.0 & 73.5 & 92.6 & 63.9 \\
 & PIM~\cite{chiaroni2023parametric} & ICCV 2023 & 94.7 & 97.4 & 93.3 & 78.3 & 84.2 & 66.5 & 83.1 & {\ul 95.3} & 77.0 \\
 & SimGCD~\cite{Wen_2023_ICCV} & ICCV 2023 & 97.1 & 95.1 & 98.1 & 80.1 & 81.2 & 77.8 & 83.0 & 93.1 & 77.9 \\
 & SPTNet~\cite{wang2024sptnet} & ICLR 2024 & 97.3 & 95.0 & \textbf{98.6} & 81.3 & 84.3 & 75.6 & \textbf{85.4} & 93.2 & \textbf{81.4} \\ 
 & ProtoGCD~\cite{10948388} & TPAMI 2025 & 97.3 & 95.3 & 98.2 & {\ul 81.9} & 82.9 & \textbf{80.0} & 84.0 & 92.2 & {\ul 79.9} \\ \bottomrule
\end{tabular}
}
\end{table*}

\begin{table*}[!t]
\setlength\tabcolsep{6pt}
\centering
\renewcommand{\arraystretch}{1}
\caption{Comparative results of several GCD methods on fine-grained datasets.}
\label{tab:gcd-fine-grained}
\resizebox{.8\linewidth}{!}{
\begin{tabular}{@{}lllcccccccccccc@{}}
\toprule
\multirow{2}{*}{Type} & \multirow{2}{*}{Methods} & \multirow{2}{*}{Venue} & \multicolumn{3}{c}{CUB} & \multicolumn{3}{c}{Stanford Cars} & \multicolumn{3}{c}{FGVC-Aircraft} & \multicolumn{3}{c}{Herbarium19} \\ \cmidrule(l){4-6} \cmidrule(l){7-9} \cmidrule(l){10-12}  \cmidrule(l){13-15} 
 &  &  & All & Old & New & All & Old & New & All & Old & New & All & Old & New \\ \midrule
\multirow{8}{*}{\begin{tabular}[c]{@{}l@{}}Non-parametric\\ Classification\end{tabular}} & K-means~\cite{arthur2007k} & Baseline  & 34.3 & 38.9 & 32.1 & 12.8 & 10.6 & 13.8 & 16.0 & 14.4 & 16.8 & 13.0 & 12.2 & 13.4 \\
 & GCD~\cite{vaze2022gcd} & CVPR 2022 & 51.3 & 56.6 & 48.7 & 39.0 & 57.6 & 29.9 & 45.0 & 41.1 & 46.9 & 35.4 & 51.0 & 27.0 \\
 & XCon~\cite{fei2022xcon} & BMVC 2022 & 52.1 & 54.3 & 51.0 & 10.5 & 58.8 & 31.7 & 47.7 & 44.4 & 49.4 & - & - & - \\
 & DCCL~\cite{pu2023dynamic} & CVPR 2023 & 63.5 & 60.8 & 64.9 & 43.1 & 55.7 & 36.2 & - & - & - & - & - & - \\
 & GPC~\cite{Zhao_2023_ICCV} & ICCV 2023 & 55.4 & 58.2 & 53.1 & 42.8 & 59.2 & 32.8 & 46.3 & 42.5 & 47.9 & 36.5 & 51.7 & 27.9 \\
 & PromptCAL~\cite{zhang2023promptcal} & CVPR 2023 & 62.9 & 64.4 & 62.1 & 50.2 & 70.1 & 40.6 & 52.2 & 52.2 & 52.3 & 37.0 & 52.0 & 28.9 \\
 & InfoSieve~\cite{rastegar2023learn} & NeurIPS 2023 & \textbf{69.4} & \textbf{77.9} & \textbf{65.2} & 55.7 & 74.8 & 46.4 & 56.3 & \textbf{63.7} & \textbf{62.5} & 41.0 & 55.4 & 33.2 \\
 & CiPR~\cite{hao2024cipr} & TMLR 2024 & 57.1 & 58.7 & 55.6 & 47.0 & 61.5 & 40.1 & - & - & - & 36.8 & 45.4 & 32.6 \\
 & CMS~\cite{choi2024contrastive} & CVPR 2024 & {\ul 68.2} & {\ul 76.5} & 64.0 & {\ul 56.9} & {\ul 76.1} & 47.6 & 56.0 & {\ul 63.4} & 52.3 & 36.4 & 54.9 & 26.4 \\ \midrule \midrule
\multirow{7}{*}{\begin{tabular}[c]{@{}l@{}}Parametric\\ Classification\end{tabular}} & RankStats+~\cite{han2021autonovel} & TPAMI 2021 & 33.3 & 51.6 & 24.2 & 28.3 & 61.8 & 12.1 & 26.9 & 36.4 & 22.2 & 27.4 & 55.8 & 12.8 \\
 & UNO+~\cite{Fini_2021_ICCV} & ICCV 2021 & 35.1 & 49.0 & 28.1 & 35.5 & 70.5 & 18.6 & 40.3 & 56.4 & 32.2 & 28.3 & 53.7 & 14.7 \\
 & ORCA~\cite{cao2022openworld} & ICLR 2022 & 35.3 & 45.6 & 30.2 & 23.5 & 50.1 & 10.7 & 22.0 & 31.8 & 17.1 & 20.9 & 30.9 & 15.5 \\
 & PIM~\cite{chiaroni2023parametric} & ICCV 2023 & 62.7 & 75.7 & 56.2 & 43.1 & 66.9 & 31.6 & - & \textbf{-} & \textbf{-} & 42.3 & 56.1 & 34.8 \\
 & SimGCD~\cite{Wen_2023_ICCV} & ICCV 2023 & 60.3 & 65.6 & 57.7 & 52.8 & 71.9 & 45.0 & 54.2 & 59.1 & 51.8 & {\ul 44.0} & 58.0 & {\ul 36.4} \\
 & $\mu$GCD~\cite{vaze2023no} & NeurIPS 2023 & 65.7 & 68.0 & 64.6 & 56.5 & 68.1 & \textbf{50.9} & 53.8 & 55.4 & 53.0 & - & - & - \\
 & SPTNet~\cite{wang2024sptnet} & ICLR 2024 & 65.8 & 68.8 & {\ul 65.1} & \textbf{59.0} & \textbf{79.2} & {\ul 49.3} & \textbf{59.3} & 61.8 & {\ul 58.1} & 43.4 & {\ul 58.7} & 35.2 \\
 & ProtoGCD~\cite{10948388} & TPAMI 2025 & 63.2 & 68.5 & 60.5 & 53.8 & 73.7 & 44.2 & {\ul 56.8} & 62.5 & 53.9 & \textbf{44.5} & \textbf{59.4} & \textbf{36.5} \\ \bottomrule
\end{tabular}
}
\end{table*}

\subsection{Evaluation of Novel Class Discovery}\label{subsec:eval}
\textbf{\emph{1) Evaluation of NCD.}}
Canonically, NCD is evaluated using two protocols: (1) task-aware, where each test sample is known to be from old or novel classes in advance, and (2) task-agnostic, where the prior information is not given and the model needs to classify the sample from all classes. The classification accuracy and cluster accuracy (see Eq.~\eqref{eq:acc} for more details) are reported for old and new classes, respectively.
The results of task-aware and task-agnostic are elaborated in Table~\ref{tab:ncd-task-aware} and Table~\ref{tab:ncd-task-agnostic}, respectively. By default, all methods use ResNet-18~\cite{he2016deep} as the backbone. All methods are evaluated on CIFAR10~\cite{krizhevsky2009learning}, CIFAR100~\cite{krizhevsky2009learning} and ImageNet~\cite{deng2009imagenet} datasets. Generally, more recent methods ComEx~\cite{yang2022divide} and IIC~\cite{Li_2023_CVPR} achieve remarkable performance.

\textbf{\emph{2) Evaluation of GCD.}}
In essence, GCD is a clustering task for both old and new categories. One could not directly compare model predictions and ground truth labels like in conventional supervised learning. In GCD, the clustering accuracy (ACC) of the model's predictions $\tilde y_i$ given the ground-truth labels $y_i$ is measured as:
\begin{equation}
    ACC=\max_{\omega\in\Omega(\mathcal{Y}_u)}\frac{1}{M}\sum_{i=1}^M\mathds{1}\big\{y_i=\omega(\tilde y_i)\big\}.
    \label{eq:acc}
\end{equation}
Here, $\mathds{1}(\cdot)$ is the indicator function, and $M$ denotes the number of test samples. $\Omega(\mathcal{Y}_u)$ represents the set of all bijective permutations that match the model predictions to the ground-truth labels. The optimal permutation in Eq.~\eqref{eq:acc} is obtained through the Hungarian algorithm~\cite{kuhn1955hungarian}, which is only implemented \emph{once} across all classes.
Here, we present a comprehensive comparative evaluation of several GCD methods, including both non-parametric and parametric types. The results on generic and fine-grained datasets are elaborated in Table~\ref{tab:gcd-generic} and Table~\ref{tab:gcd-fine-grained}, respectively. By default, the backbone selects DINO~\cite{caron2021emerging} pre-trained ViT-B/16~\cite{dosovitskiy2021an}, and only the last transformer block is tuned. The $\texttt{[CLS]}$ token of the last layer is selected as the feature representation. The evaluation datasets include generic image classification datasets in Table~\ref{tab:gcd-generic}, \eg, CIFAR10~\cite{krizhevsky2009learning}, CIFAR100~\cite{krizhevsky2009learning} and ImageNet~\cite{deng2009imagenet} subsets, as well as fine-grained datasets in Table~\ref{tab:gcd-fine-grained}, like CUB~\cite{wah2011caltech}, Stanford Cars~\cite{krause20133d}, FGVC-Aircraft~\cite{maji2013fine} and Herbarium 19~\cite{tan2019herbarium}. By default, 50\% of the classes serve as the old classes, while others serve as novel classes. Overall, the recent parametric classification methods~\cite{vaze2023no,wang2024sptnet,10948388} achieve better results.

\begin{figure*}[t]
  \begin{center}
\centerline{\includegraphics[width=\textwidth]{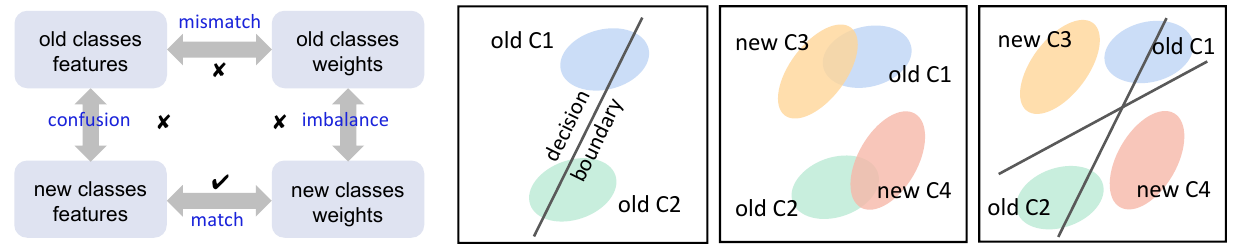}}
   \caption{(a) Updating the model continuously on new classes without accessing old data leads to three phenomena: (b) (old-old) representation and classifier mismatch, (c) (old-new) representation confusion, and (d) (old-new) classifier imbalance. Those three issues result in the catastrophic forgetting problem.}
   \label{fig:cil-problem-analysis}
 \end{center}
 \vskip -0.05in
\end{figure*}
\section{Continual Learning}
\label{sec:class-incremental-learning}

After detecting unknown samples via OOD detection or open-set recognition techniques (Sec.~\ref{sec:unknown-rejection}), those samples should be labeled by humans or novel class discovery strategies (Sec.~\ref{sec:novel-discovery}). Then, the system must continually extend the multi-class classifier to learn those new classes, which is referred to as continual learning (CL), being the third step in the open-world recognition process (Fig.~\ref{figure-2}).
In this section, we provide a comprehensive survey on recent advances in CL. To be specific, we first present the problem definition and key challenges in Sec.~\ref{sec:problem-formulation}. Then, we group existing CL studies into several main categories: regularization methods (Sec.~\ref{sec:regularization}), data replay methods (Sec.~\ref{sec:data-replay}), feature replay methods (Sec.~\ref{sec:feature-replay}) and model expansion methods (Sec.~\ref{sec:model-expansion}). Then, we demonstrate some recent extended practical settings of CL (Sec.~\ref{subsec:cil-extend}) and theoretical advance of continual learning (Sec.~\ref{subsec:cl-theory}).
Finally, some evaluation results are presented (Sec.~\ref{subsec:cl-evaluation}).

\subsection{Problem Formulation and Key Challenges}
\label{sec:problem-formulation}

\emph{\textbf{1) Problem Formulation.}} 
Typically, a CL problem involves the sequential learning of tasks that consist of disjoint class sets, and the model has to learn a unified classifier that can classify all seen classes. At incremental step $t$, a dataset $\mathcal{D}_{t} = \{\bm{x}_{i}^{t}, y_{i}^{t}\}^{n_{t}}_{i=1}$ is given, where $\bm{x}$ is an image in the input space $\mathcal{X}$ and $y \in \mathcal{C}_{t}$ is its corresponding label. $\mathcal{C}_{t}$ is the class set of task $t$ and the class sets of different task are disjoint, \ie, $\mathcal{C}_{i} \cap \mathcal{C}_{j} = \emptyset$ if $i \neq j$.
To facilitate analysis, we represent the DNN-based model with two components: a feature extractor and a unified classifier. Specifically, the feature extractor $f_{\bm{\theta}}: \mathcal{X} \rightarrow \mathcal{Z}$, 
parameterized by $\bm{\theta}$, maps the input $\bm{x}$ into a feature vector $\bm{z} = f_{\bm{\theta}}(\bm{x}) \in \mathbb{R}^{d}$ in the feature space $\mathcal{Z}$; the classifier $g_{\bm{\varphi}}: \mathcal{Z} \rightarrow \mathbb{R}^{|\mathcal{C}_{1:t}|}$, parameterized by $\bm{\varphi}$, produces a probability distribution $g_{\bm{\varphi}}(\bm{z})$ as the prediction for $\bm{x}$. Denote the overall parameters by $\bm{\Theta} = (\bm{\theta}, \bm{\varphi})$.
At stage $t$, the general objective is to minimize a predefined loss function $\ell$ (\ie, cross-entropy loss) on the new dataset $\mathcal{D}_{t}$ without interfering and with possibly improving on those that were learned previously~\cite{Aljundi2019ContinualLI}: 
\begin{equation}\label{eqcil}
\begin{aligned}
\argmin\limits_{\bm{\theta},\bm{\varphi},\epsilon} \mathbb{E}_{(\bm{x},y) \backsim \mathcal{D}_{t}}[\ell(&g_{\bm{\varphi}}(f_{\bm{\theta}}(\bm{x})), y)] + \sum\nolimits \epsilon_{i}  \\
	\text{s.t.}~~ \mathbb{E}_{(\bm{x},y) \backsim \mathcal{D}_{i}}[\ell(g_{\bm{\varphi}}(f_{\bm{\theta}}(\bm{x})), y) &-\ell(g_{\bm{\varphi}^{t-1}}(f_{\bm{\theta}^{t-1}}(\bm{x})), y)] \leqslant \epsilon_{i}, \\ \epsilon_{i} \geqslant 0; ~~ \forall &i \in [1,t-1].
\end{aligned}
\end{equation}
The last term $\epsilon = \{\epsilon_{i}\}$ is a slack variable that tolerates a small increase in the loss on datasets of old tasks.

\emph{\textbf{2) Key Challenges.}} The central challenge of CL is that old data are unavailable when learning new classes and the objective defined in Eq.~\ref{eqcil} can not be optimized directly. Therefore, the best configuration of the model for all seen classes must be sought by minimizing $\mathcal{L}_{t}$ on $\mathcal{D}_{t}$:
\begin{equation}\label{eq7}
\begin{aligned}
\mathcal{L}_{t} = \mathbb{E}_{(\bm{x},y) \backsim \mathcal{D}_{t}}[\ell(g_{\bm{\varphi}}(f_{\bm{\theta}}(\bm{x})), y)].
\end{aligned}
\end{equation}
However, directly updating the model on new classes would cause the catastrophic forgetting problem \cite{Goodfellow2014AnEI, McCloskey1989CatastrophicII, french1995interactive, ni2024enhancing}, \ie, the classification accuracy on old classes drops dramatically after learning new classes. By taking a closer look at catastrophic forgetting in CL, we find that there are mainly three issues, and each of them represents one aspect of forgetting, while the concurrence of them results in the performance degradation of old classes after learning new classes in CL, as demonstrated as follows and illustrated in Fig.~\ref{fig:cil-problem-analysis}. 
(1) \emph{Representation and classifier mismatch}: Without accessing old training data, directly updating the model on new classes could dramatically change the learned representations and class weights for old classes, which leads to the mismatch between old class features and old class weights. Consequently, the input-output behavior of old classes would be forgotten.
(2) \emph{Representation confusion}: After updating the model on new classes, the feature extractor would no longer be suitable for old classes. As a result, the input from old and new classes would be confused and overlapped in the deep feature space, making it hard to distinguish them at inference time.
(2) \emph{Classifier imbalance}: When learning new classes, the class weights of old and new classes can not be jointly optimized because of the lack of old data. Therefore, the class weights of old and new classes are often imbalanced, and the decision boundary can easily shift toward novel classes. The above analysis would not only give a deeper understanding of catastrophic forgetting in CL but also provide a unified perspective to understand existing methods.

\subsection{Regularization Methods}
\label{sec:regularization}
Both the parameters of a model and its input-output behavior provide distinct measures of what the model has learned from different perspectives. As a result, methods in this domain can be categorized into two groups: explicit and implicit regularization, each aimed at preserving specific forms of knowledge. Explicit regularization seeks to prevent substantial alteration of crucial network parameters, while implicit regularization focuses on maintaining the input-output behavior of the network.

\emph{\textbf{1) Explicit Weight Constraint.}} For explicit regularization, the central and challenging work is to estimate the importance of network parameters. After learning each task, the importance of individual parameters in the model is estimated to regularize the new class learning process by penalizing the change of important parameters. The general regularization loss can be formalized as:
\begin{equation}\label{eq8}
	\mathcal{L}_{\text{E-reg}, t} = \sum_{i}^{P}\Omega_{i}(\bm{\Theta}_{i}^{t}-\bm{\Theta}_{i}^{t-1})^2,
\end{equation}
where $\Omega_{i}$ is the estimated importance factor for parameter $\bm{\Theta}_{i}$, and $P$ is the number of parameters in the network.
As can be seen, the central and challenging work is to estimate the importance of network parameters.
The first work to achieve this goal is elastic weight consolidation (EWC) \cite{Kirkpatrick2017OvercomingCF}, which uses the diagonal of the Fisher information matrix to compute the importance $\Omega_{i}$. Alternatively, parameter importance can be estimated by quantifying the sensitivity of the predicted output function to changes in each parameter \cite{aljundi2018memory}. Kong et al., \cite{kong2023overcoming} theoretically shown that the forgetting has a relationship with the maximum eigenvalue of the Hessian matrix, and proposed to explore the maximum eigenvalue to overcome catastrophic forgetting. 
Following those methods, various weight constraint approaches have been developed, such as RWalk \cite{chaudhry2018riemannian}, VCL \cite{nguyen2018variational}, UCL \cite{ahn2019uncertainty}, KCL \cite{derakhshani2021kernel} and AGS-CL \cite{jung2020continual}.
These methods can be interpreted as metaplasticity-inspired \cite{abraham1996metaplasticity} from a neuroscience point of view \cite{van2020brain}.
Nonetheless, designing a reasonable importance metric within a neural network remains challenging. 
Another line of works such as OWM \cite{zeng2019continual}, OGD \cite{farajtabar2020orthogonal}, AOP \cite{guo2022adaptive}, GPM \cite{sahagradient}, Adam-NSCL \cite{wang2021training}, NCL \cite{kao2021natural} and DSFGP \cite{yang2025revisiting} avoid interfering with previously learned knowledge from a parameter space perspective by projecting and updating the parameters in the orthogonal direction or the null space of the previous tasks. Those methods assume that the capacity of a neural network is high enough to learn new tasks effectively, which would not always hold in practice, especially for continual learning with long sequences.

\begin{figure*}[t]
  \begin{center}
\centerline{\includegraphics[width=\textwidth]{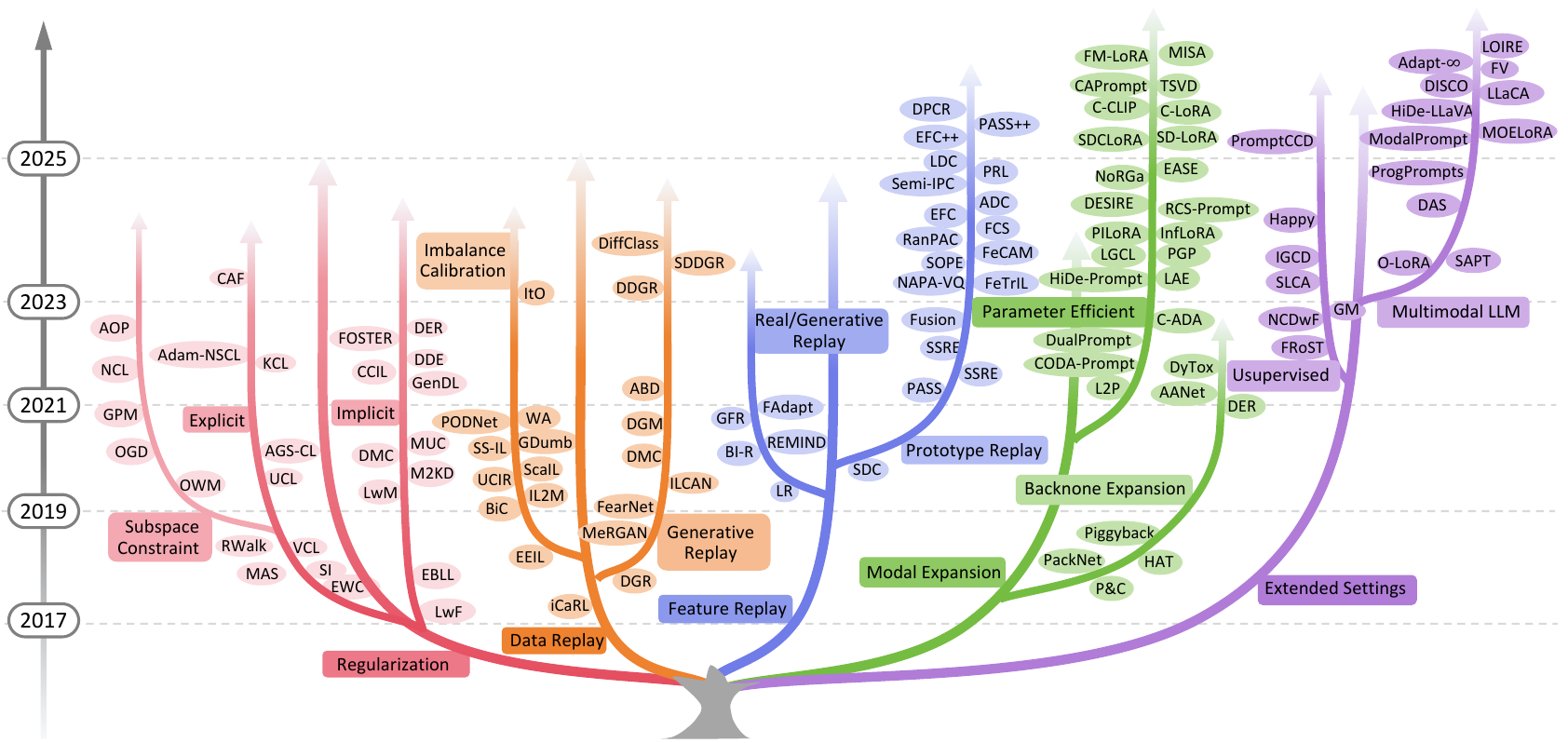}}
   \caption{The evolutionary tree of representative continual learning methods.}
   \label{fig:tree-cil}
 \end{center}
 \vskip -0.1in
\end{figure*}

\emph{\textbf{2) Implicit Knowledge Distillation.}} Rather than directly constraining the parameters of the network, implicit regularization approaches focus on keeping the input-output behavior of the network based on current training data. The general regularization loss can be formalized as:
\begin{equation}\label{eq9}
\mathcal{L}_{\text{I-reg}, t} = \mathbb{E}_{(\bm{x},y) \backsim \mathcal{D}_{t}}[D(\text{out}^{t-1}(\bm{x}), \text{out}^{t}(\bm{x}))],
\end{equation}
where $D(\cdot)$ represents the distance metric such as Euclidean distance, $cos$ distance and Kullback-Leibler (KL) divergence.
To realize this idea, Li \etal~\cite{Li2018LearningWF} proposed learning without forgetting (LwF), which firstly incorporates the knowledge distillation (KD) \cite{hinton2015distilling} technique into continual learning to consolidate old knowledge by distilling the outputs of the previous model to that of the updated model as follows:
\begin{equation}\label{eq10}
	\mathcal{L}_{\text{LwF}, t} = \mathbb{E}_{(\bm{x},y) \backsim \mathcal{D}_{t}}\left[-\sum_{i=1}^{|\mathcal{C}_{\text{old}}|} p_{i}^{t-1}(\bm{x}) \text{log}~p_{i}^{t}(\bm{x})\right],
\end{equation}
where $|\mathcal{C}_{\text{old}}| = |\mathcal{C}_{1:t-1}|$ is the number of old classes and $p_{i}^{t-1}(\bm{x})$, $p_{i}^{t}(\bm{x})$ are the previous and current probabilities after temperature scaling. By optimizing $\mathcal{L}_{LwF, t}$, the KL divergence between the two probability distributions could be minimized. Noteworthy, although KD \cite{hinton2015distilling} was originally developed for model compression, it has been widely used in CL and served as a basic building block in many methods. 
\begin{figure}[!t]
\centering
\vskip 0.03in
\includegraphics[width=\columnwidth]{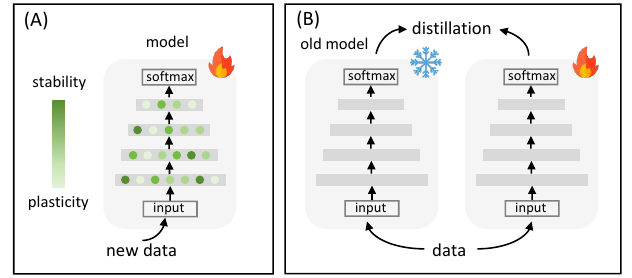}
\vskip -0.07in
\caption{Illustration of regularization methods. (A) Explicit weight constraint. (B) Implicit knowledge distillation.}
\label{Figure:cil_reg}
\vskip -0.1in
\end{figure}

Later works followed the teacher-student distillation framework to address the catastrophic forgetting in CL by modifying the KD technique. Some approaches distill important features in the deep feature space \cite{rannen2017encoder, DharSPWC19}, where important features can be identified via attention mechanisms \cite{DharSPWC19, zagoruyko2016paying} or feature projection \cite{rannen2017encoder}. For example, to evaluate the importance of features, encoder-based lifelong learning (EBLL) \cite{rannen2017encoder} projects the feature to a subspace with lower dimensions based on autoencoders. Then the Euclidean distance is used to distill those important features. Learning without memorization (LwM) \cite{DharSPWC19} identifies the important features based on attention technique \cite{zagoruyko2016paying}, and then penalizes the changes in classifiers' attention maps to retain important information of the old classes when learning new classes. By distilling knowledge from important features, the model not only preserves the crucial information of old classes but also gives more
space to adjust to the new classes. Other works explore methods to distillate knowledge from multi-models \cite{zhang2020class, liu2020more} or auxiliary data \cite{zhang2020class, lee2019overcoming, zhu2021calibration}. For instance, deep model consolidation (DMC) \cite{zhang2020class} distills the knowledge from both the old model and the new model via a novel double distillation training objective. Instead of sequentially distilling knowledge only from the penultimate model, M2KD \cite{zhou2019m2kd} directly leverages all previous models for KD. Alternatively, MUC \cite{liu2020more} distills knowledge from an ensemble of auxiliary classifiers, to estimate more efficient and effective regularization constraints. Besides, some studies found that distillation on new classes can not retain old knowledge effectively, and the model could easily be biased towards new classes. To address this problem, Lee et al., \cite{lee2019overcoming} leverages a large stream of unlabeled auxiliary data for KD. CCIL \etal~\cite{zhu2021calibration} adopts cutout \cite{devries2017improved} to synthesize novel data for KD, and demonstrated its effectiveness on calibrating the model's predictions. However, as noticed by previous works \cite{Hsu2018ReevaluatingCL, Ven2019ThreeSF, van2020brain}, those regularization methods often failed dramatically for class continual learning scenario. Therefore, KD technique is often combined with data replay \cite{Hou2019LearningAU, Douillard2020SmallTaskIL, buzzega2020dark, simon2021learning, hu2021distilling, wang2022foster}, which will be detailed below.

\subsection{Data Replay Methods}
\label{sec:data-replay}
A basic assumption in CL is the inaccessibility of old training data. However, data replay methods relax this constraint by saving and replaying some training samples of old classes,\ie, $\mathcal{M}$. Then the model is jointly trained on $\mathcal{M} \bigcup \mathcal{D}_\text{t}$. Intuitively, with replayed data, the previous knowledge would be consolidated, and the three problems (demonstrated in Fig.~\ref{fig:cil-problem-analysis}) in CL would be alleviated: First, the representation and classifier of old classes would be better matched because old samples are relearned in an end to end manner when learning new classes. Second, the representation confusion and class weights imbalance problems between old and new classes would be alleviated because data replay allows the discrimination between old and new classes.
Rebuffi \etal~\cite{Rebuffi2017iCaRLIC} firstly investigated the data replay strategy for CL. Due to the strong performance, various data replay methods have been developed, mainly focusing on imbalance calibration and generative replay.

\emph{\textbf{1) Imbalance Calibration.}} Due to the memory limitation, the number of saved samples is usually much lower compared with that of new classes. Therefore, there exists the class imbalance problem \cite{zhang2023towards} between new and old classes. Visually, the norms of the weight vectors of old classes are much smaller than those of new classes \cite{ZhaoXGZX20}. To calibrate the bias, some approaches aim to learn a balanced classifier during training, while other approaches use post-hoc methods to calibrate the classifier. 
A simple and classical way is under-sampling \cite{lemaitre2017imbalanced, prabhu2020greedy}. For example, end-to-end incremental learning (EEIL) \cite{castro2018end} includes additional balanced fine-tuning at the end of each incremental stage. Specifically, a balanced dataset that contains the same number of samples per class is constructed by reducing the number of samples from the new classes. Then, the balanced dataset is used to fine-tune the model with a small learning rate. Lee et al., \cite{lee2019overcoming} also adopt a balanced fine-tuning strategy by scaling the gradient based on the number of samples in each class. GDumb \cite{prabhu2020greedy} greedily stores samples in memory and trains a model from scratch using samples only in the memory at inference time. These methods could alleviate the forgetting problem in CL, but lead to under-fitting of new classes since a lot of samples of new classes are not learned.

Other training-time imbalance calibration methods explore modifying the classifier directly, such as cosine normalization \cite{Hou2019LearningAU} and separated softmax \cite{ahn2021ss, zhu2021calibration}. A representative method is UCIR \cite{Hou2019LearningAU}, which adopts cosine normalization and margin ranking loss to learn a balanced classifier. On the one hand, UCIR normalizes both the input and class weights in the softmax options when computing the predicted probability of a sample:
\begin{equation}
p(c|\bm{x}) = \frac{\text{exp}{(\eta\langle\bar{f}_{\bm{\theta}}(\bm{x}), \bar{\bm{\varphi}}_c}\rangle)}{\sum_{j} \text{exp}{(\eta\langle\bar{f}_{\bm{\theta}}(\bm{x}), \bar{\bm{\varphi}}_j}\rangle)},
\end{equation}
where $\bar{f}_{\bm{\theta}}(\cdot) = \frac{f_{\bm{\theta}}(\cdot)}{\Vert f_{\bm{\theta}}(\cdot) \Vert}$ and $\bar{\bm{\varphi}} = \frac{\bm{\varphi}}{\Vert \bm{\varphi} \Vert}$ denotes the normalized vector. $\langle \cdot , \cdot \rangle$ is the inner product, and $\eta$ is a temperature parameter controlling the hardness of the softmax distribution.
On the other hand, UCIR introduces a margin ranking loss for reserved old samples $\bm{x} \in \mathcal{M}_{\text{old}}$ to separate the new classes from old classes in the deep feature space:
\begin{equation}
\mathcal{L}_{mr, t}(\bm{x}) = \sum_{k=1}^{K} \text{max}~ (m - \langle\bar{f}_{\bm{\theta}}(\bm{x}), \bar{\bm{\varphi}}_y\rangle + \langle\bar{f}_{\bm{\theta}}(\bm{x}), \bar{\bm{\varphi}}_k\rangle),
\end{equation}
where $m$ is the margin, $\bar{\bm{\varphi}}_y$ and $\bar{\bm{\varphi}}_k$ refer to the ground truth and the top-K new class embeddings, respectively.
Those two strategies can effectively eliminate the classifier bias in CL.
Ahn \etal~\cite{ahn2021ss} systematically analyzed the cause of classifier bias, and proposed a method that integrates separated softmax and task-wise knowledge distillation to resolve the bias. Those two strategies alleviate the bias accumulation in the CL process. Recently, Liu \etal~\cite{liu2021rmm} proposed a reinforcement learning based strategy to learn both the task-level and class-level memory allocation in different incremental phases to address the data imbalance problem.

\begin{figure}[t]
\centering
\vskip 0.03in
\includegraphics[width=\columnwidth]{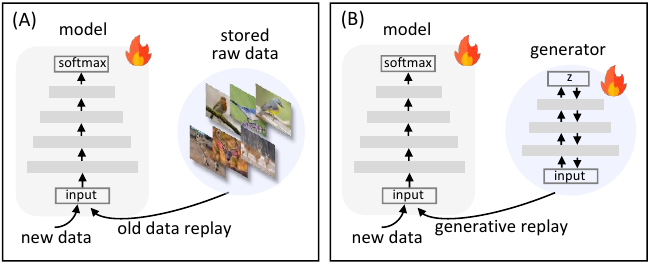}
\vskip -0.05in
\caption{Illustration of data replay methods. (A) Real data replay. (B) Generative data replay.}
\label{Figure:cil_dr}
\end{figure}

\begin{table*}[!t]
    \centering
    \caption{Summarization of imbalance calibration strategies.}
    \vskip -0.05in
    \resizebox{\textwidth}{!}{
    \begin{tabular}{lccccc}
        \toprule
        \multirow{2}{*}{Methods} & \multicolumn{4}{c}{Imbalance Calibration Settings} & \multirow{2}{*}{Imbalance Calibration Strategy} \\
        \cmidrule{2-3} \cmidrule{4-5}
         & Data & Classifier & Train & Test & \\
        \midrule
        
        \rowcolor[rgb]{ .949,  .949,  .949} 
        EEIL~\cite{castro2018end}       & \textcolor{green!50!black}{\ding{52}} & & \textcolor{red}{\ding{52}} & & Two-stage: construct a balanced dataset to fine-tune the model. \\

        GDumb~\cite{prabhu2020greedy}   & \textcolor{green!50!black}{\ding{52}} & & \textcolor{red}{\ding{52}} & & Down-sampling: build a balanced dataset and train the model from scratch. \\

        \rowcolor[rgb]{ .949,  .949,  .949} 
        SS-IL~\cite{ahn2021ss}         & & \textcolor{green!50!black}{\ding{52}} & \textcolor{red}{\ding{52}} & & Decouple softmax operations and apply knowledge distillation. \\

        RMM~\cite{liu2021rmm}    & \textcolor{green!50!black}{\ding{52}} & & \textcolor{red}{\ding{52}} & & Reinforcement learning to optimize class balance in memory. \\

        \rowcolor[rgb]{ .949,  .949,  .949} 
        UCIR~\cite{Hou2019LearningAU}        & & \textcolor{green!50!black}{\ding{52}} & \textcolor{red}{\ding{52}} & & Feature and weight normalization with ranking loss. \\

        iCaRL~\cite{Rebuffi2017iCaRLIC}  & & \textcolor{green!50!black}{\ding{52}} & & \textcolor{red}{\ding{52}} & Classify based on nearest class mean. \\

        \rowcolor[rgb]{ .949,  .949,  .949} 
        BiC~\cite{Wu2019LargeSI}         & & \textcolor{green!50!black}{\ding{52}} & & \textcolor{red}{\ding{52}} & Calibrate prediction bias using a separate balanced validation set. \\

        WA~\cite{ZhaoXGZX20}  & & \textcolor{green!50!black}{\ding{52}} & & \textcolor{red}{\ding{52}} & Align magnitudes of classification weights between new and old classes. \\

        \rowcolor[rgb]{ .949,  .949,  .949} 
        IL2M~\cite{belouadah2019il2m}  & & \textcolor{green!50!black}{\ding{52}} & & \textcolor{red}{\ding{52}} & Post-hoc calibration of model's probability distribution. \\

        ScaIL~\cite{belouadah2020scail}  & & \textcolor{green!50!black}{\ding{52}} & & \textcolor{red}{\ding{52}} & Reshape classifier using aggregate class statistics. \\

        \rowcolor[rgb]{ .949,  .949,  .949} 
        ItO~\cite{zhu2023imitating}  & & \textcolor{green!50!black}{\ding{52}} &\textcolor{red}{\ding{52}} &  & Introduce deviation compensation to maintain learned decision boundary. \\
        \bottomrule
    \end{tabular}}
    \label{tab:bias_calibration_options}
\end{table*}
For post-hoc imbalance calibration, Rebuffi \etal~\cite{Rebuffi2017iCaRLIC} notice the classifier imbalance problem and introduce the nearest class mean (NCM) classifier at inference time. Technically, the NCM is computed by averaging the feature representations of each class:
\begin{equation}
	\bm{\varphi}_{c} \leftarrow \frac{1}{n_c}\sum_i \mathbb I (y_i = c) f_{\bm{\theta}}(\bm{x}_i), ~~ \forall c \in \mathcal{C},
\end{equation}
where $\mathcal{C}$ denotes all the encountered classes, $n_c$ denotes the number of the samples in class $c$. Particularly, for old classes, only the preserved data are used to compute the class mean. Compared with the original linear classifier, the NCM classifier is less sensitive to the class imbalance problem~\cite{yang2020convolutional}. A more direct way is to calibrate the classifier via bias correction layer~\cite{Wu2019LargeSI} or output scaling \cite{ZhaoXGZX20, belouadah2020scail, belouadah2019il2m}. For instance, Bias
Correction (BiC)~\cite{Wu2019LargeSI} calibrates the output logits of the model by learning a linear bias correction layer:
\begin{equation}
	o'_{c} \leftarrow \alpha o_{c} + \beta, ~~ \forall c \in \mathcal{C}_{\text{new}},
\end{equation}
where $o_{c}$ is the output logits for the new classes $c \in \mathcal{C}_{\text{new}}$. The bias calibration parameters \emph{i.e} $\alpha$ and $\beta$ are estimated based on a balanced validation set and shared by all new classes.
Weight Alignment (WA) \cite{ZhaoXGZX20} scales the output logits as follows:
\begin{equation}
	o'_{c} \leftarrow \gamma o_{c}, ~~ \forall c \in \mathcal{C}_{\text{new}},
\end{equation}
where
\begin{equation}
	\gamma = \frac{\text{Mean}((\Vert \bar{\bm{\varphi}}_1 \Vert, ..., \Vert \bar{\bm{\varphi}}_{|\mathcal{C}_{\text{old}}|} \Vert))}{\text{Mean}((\Vert \bar{\bm{\varphi}}_{|\mathcal{C}_{\text{old}}|+1} \Vert, ..., \Vert \bar{\bm{\varphi}}_{|\mathcal{C}_{\text{old}}|+|\mathcal{C}_{\text{new}}|} \Vert))},
\end{equation}
in which Mean($\cdot$) returns the mean value of elements in the norm of
classifier weights. Since scaling output logits is equivalent to scaling the norm of the weights, the method can be viewed as a strategy to align the average norm of the weight vectors for old and new classes.
Similarly, ScaIL \cite{belouadah2020scail} also scales the classifier weights by leveraging statistical information of the model outputs. 
Instead of scaling the classifier weights or output logits, Belouadah \etal~\cite{belouadah2019il2m} rectified the output softmax probability by utilizing class statistics, \ie, the prediction score of the classifier. Concretely, if the prediction of an input $\bm{x}$ belongs to new classes, the softmax probability on any old class $c \in \mathcal{C}_{\text{old}}$ would be modified as follows:
\begin{equation}
	p'(c|\bm{x}) = p(c|\bm{x}) \cdot \frac{\mu^P (c)}{\mu^N (c)} \cdot \frac{\mu^N (\mathcal{C}_{\text{new}})}{\mu^P (\mathcal{C}_{\text{old}})}, \quad \forall c \in \mathcal{C}_{\text{old}},
\end{equation}
where $P$ denotes the state where the old class $c$ was learned initially; $N$ represents the current incremental state. $\mu^P (c)$ represents the mean prediction
scores of all training samples of class $c$ at state P, and $\mu^N (c)$ represents the mean prediction
scores of preserved samples of class $c$ at current state N. $\mu^P (\mathcal{C}_{\text{old}})$ and $\mu^N (\mathcal{C}_{\text{new}})$ share the similar meanings for all old classes $\mathcal{C}_{\text{ole}}$ and new classes $\mathcal{C}_{\text{new}}$, respectively. Table \ref{tab:bias_calibration_options} provides a summary and comparison of different imbalance strategies in continual learning. Recently, Lai \etal~\cite{lai2025pareto} proposed a novel Pareto continual learning that reformulates the stability-plasticity trade-off in data replay based continual learning as a multi-objective optimization problem.
Xu \etal~\cite{xu2025practical} revealed the ineffectiveness of existing popular data replay methods and called for a more flexible, and practical CL setting where the storage buffer can be dynamically adjusted according to acceptable cost.

\emph{\textbf{2) Generative Data Replay.}} Despite the success of real data replay in alleviating catastrophic forgetting, there are issues such as violation of data privacy and memory limitation: storing a fraction of old raw data would be unsuitable for privacy-sensitive applications in the field of healthcare and security \cite{li2020federated, trepte2021social}, and suffer from memory limitation for long-step continual learning and applications with limited computational budget. Moreover, directly storing some old training samples is less human-like from the biological perspective \cite{kitamura2017engrams, kudithipudi2022biological, zeng2019continual}. To address privacy and memory limitation concerns of real data replay, researchers have started to consider generative data replay.
Early approaches \cite{shin2017continual, wu2018memory, xiang2019incremental} leverage the generative model (\emph{\eg}, GAN \cite{goodfellow2014generative, OdenaOS17} and autoencoder \cite{autoencoder_paper}) to generate pseudo-samples for old classes. Recent investigations explore newly developed diffusion models \cite{yang2023diffusion} for replay, \cite{gao2023ddgr, kim2024sddgr, meng2024diffclass}, enabling bidirectional instruction between generator and classifier to mitigate forgetting.
Particularly, the generative model is trained simultaneously with the classification model at each incremental stage. However, those approaches perform poorly for CL, and the generated pseudo-examples rely heavily on the quality of the generative model. Moreover, it is computational and memory-intensive to train generative models, and they also suffer from catastrophic forgetting during continual learning. 
Rather than leveraging additional complex generative models, some recent works \cite{yin2020dreaming, smith2021always, gao2022r} explore to use of the classification model itself to generate pseudo-samples. Yin \etal propose the DeepInversion \cite{yin2020dreaming} to synthesize and replay pseudo-samples of previous classes when learning new classes. However, as noticed by \cite{smith2021always, gao2022r}, the feature distribution of synthetic old samples is severely mismatched with that of real old samples, which introduces bias and misleads the decision boundary between old and new classes. Always Be Dreaming (ABD) \cite{smith2021always} mitigates this issue by using separated softmax and balanced finetuning. Gao et al.,  \cite{gao2022r} further improve ABD by leveraging relational knowledge distillation \cite{park2019relational}, which reduces forgetting while improving plasticity.

\subsection{Feature Replay Methods}
\label{sec:feature-replay}
Data replay methods suffer from memory limitations and privacy concerns in real-world applications, and generating exemplars in input space is quite challenging and highly dependent on the quality of generative models. To avoid directly storing and replaying old training data, feature replay strategies have been explored recently. Among them, some methods store and replay feature embeddings in the deep feature space, others only store the prototype, \ie, the class mean of each old class. Moreover, some works developed a generative feature replay technique without strong real feature embeddings. Compared with raw data replay, feature replay has advantages in terms of computation efficiency, memory footprint, and privacy safety. However, with the continual updating of the feature extractor in the continual learning process, the saved old features would be less valid and representative. Therefore, maintaining the effectiveness of saved features is a key challenge for feature replay methods.

\emph{\textbf{1) Real and Generative Feature Replay.}}
In DNNs, there are many features with different levels of depth or semantics. Therefore, both feature instances in intermediate \cite{pellegrini2020latent, hayes2020remind} or final layer \cite{iscen2020memory} can be saved. For feature replay in the intermediate layer, to keep the stored features valid and stable when learning new classes, one can directly freeze the layers between input and feature replay layer \cite{hayes2020remind}, or assign a lower learning rate for those layers \cite{pellegrini2020latent}. For feature replay in the final layer, freezing all layers or assigning a lower learning rate would make the model intractable to learn new knowledge. To keep the stored features valid, a feature adaptation module can be constructed to map the saved features to the updated feature space \cite{iscen2020memory}, and the classifier can then be jointly optimized over new and old classes in the same feature space. 
Besides, inspired by generative exemplar replay, some works \cite{xiang2019incremental, shen2021generative, liu2020generative} explored generative feature replay, which generates pseudo features of old classes for replay when learning new classes. For example, Xiang \etal~\cite{xiang2019incremental} employ an adversarial training scheme to produce mid‑level convolutional features whose distribution closely matches that of old classes. Liu \etal~\cite{liu2020generative} systematically show that synthesizing embeddings in early network layers incurs severe performance degradation, while generating the final features has been shown to be more effective. More recently, van de Ven et al., \cite{van2020brain} introduce a brain‑inspired feature replay mechanism that uses the classifier itself to recreate embeddings of old classes, demonstrating strong performance.

\begin{figure}[!t]
\centering
\includegraphics[width=\columnwidth]{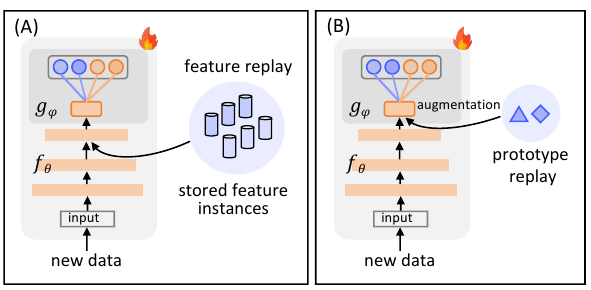}
\vskip -0.05in
\caption{Illustration of feature replay methods. (A) Real or generative feature instances replay. (B) Prototype replay.}
\label{Figure:cil_fr}
\end{figure}
\emph{\textbf{2) Prototype Replay.}}
To further reduce the memory cost of feature replay, prototype replay methods \cite{zhu2021prototype, zhu2022self, yu2020semantic} only memorize one class-representative prototype (class mean in deep feature space) for each old class. Prototype Augmentation and Self‑Supervision (PASS) memorizes a single class‑representative prototype (the class mean) for each old class and applies simple Gaussian‑noise‑based augmentation to preserve the decision boundaries of previous tasks. Specifically, for each old class $k$, pseudo feature instances are generated by augmenting the corresponding prototype:
	\begin{equation}
		\label{eq5}
		\begin{aligned}
			\widetilde{\bm{z}}_{k} = \bm{\mu}_{k} + r \cdot \bm{e},
		\end{aligned}
	\end{equation}
where $\bm{e} \sim \mathcal{N}(\bm{0}, 1)$ is Gaussian noise with the same dimensionality as the prototype $\bm{\mu}_{k} = \frac{1}{n_{k}}\sum\nolimits_{j=1}^{n_{k}}f_{\bm{\theta}}(\bm{x}_{j})$. Hyperparameter $r$ is a scaling factor that controls the level of uncertainty of the augmented features. A key point in the prototype replay is that the saved prototypes become increasingly outdated when updating the model on new classes. To keep stored prototypes valid and representations stable, PASS \cite{zhu2021prototype} incorporates self‑supervised learning to obtain more generalizable and transferable features, which in turn reduces feature drift when learning new classes.  Other works explicitly model and constrain the feature drifts of old classes via linear \cite{yu2020semantic} or nonlinear transformations \cite{toldo2022bring}. For instance, Toldo et al., \cite{toldo2022bring} model both semantic drift, by learning relationships between novel and existing prototypes, and feature drift, by tracking the evolution of new data embeddings. They jointly exploit these drift estimates to update saved prototypes.
In addition to saving only the prototypes of old classes, more distribution information can be beneficial for maintaining the decision boundary of previous tasks, and an infinite number of pseudo-feature instances can be generated implicitly \cite{zhu2021class}. Some works \cite{roy2022class,NEURIPS2024_5ae0f7cf, liu2024towards,liu2025class} directly explore learning discriminative prototypes for continual learning. Recent prototype replay methods freeze the backbone and learn a prototype-based classifier. For instance, FeCAM \cite{goswami2023fecam} leverages a frozen feature extractor and anisotropic Mahalanobis distance to address heterogeneous feature distributions, while FeTrIL \cite{petit2023fetril} builds a LinearSVCs classifier \cite{pedregosa2011scikit} on the frozen backbone. RanPAC \cite{mcdonnell2023ranpac} and ACIL \cite{zhuang2022acil} transform prototypes of old classes with random projection and then construct a linear regression classifier for all learned classes. Such methods can also be applied to pre-trained models. In general, prototype replay is simple and memory efficient, yet still effective for forgetting mitigation in continual learning.

\subsection{Model Expansion} \label{sec:model-expansion}
\emph{\textbf{1) Backbone Expansion.}}
In the task continual learning setting, dynamically extending the network architecture has been demonstrated to be quite effective \cite{mallya2018packnet, serra2018overcoming}. Specifically, when learning a new task, the old parameters are frozen, and new branches are allocated to learn new tasks. 
However, those methods are often impractical for CL because the task identity is not available at inference time, and one can not know which group of parameters should be used for inference given an input. Recently, based on exemplar replay, several architecture-based methods have been proposed for CL. Adaptive Aggregation Networks (AANets) \cite{liu2021adaptive} explores a novel network architecture to explicitly address the stability and plasticity dilemma in class continual learning. Specifically, two types of residual blocks with different numbers of learnable parameters are designed: a stable block to preserve knowledge of old classes, and a plastic block to acquire new classes. Their output feature maps are aggregated before entering the next residual level. A key step in AANets is learning the aggregation weights using a balanced mix of old and new data. A similar idea has been explored in~\cite{pham2021dualnet}, which designed a dual network architecture that includes a slow learning module for generic representation learning and a fast learning module for task-specific representation learning. The slow system is trained via contrastive learning, while the fast system is trained via supervised learning. Dynamically Expandable Representation (DER) \cite{yan2021dynamically} dynamically expands the feature extractor to retain old knowledge while learning new concepts. At each incremental stage, the existing feature extractor is frozen, and a new extractor is introduced for the new classes. However, incrementally creating new models leads to an increase in the number of parameters, while KD can be an effective way to remove redundant parameters \cite{wang2022foster}.

\begin{figure*}[t]
  \begin{center}
\centerline{\includegraphics[width=\textwidth]{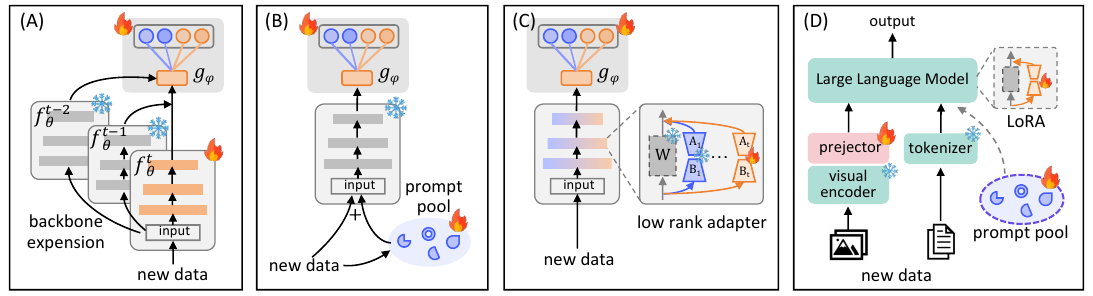}}
\caption{Illustration of model expansion methods. (A) Backbone expansion. (B) Prompt-based expansion. (C) LoRA-based expansion. (D) Parameter-efficient expansion for continual learning of multimodal large language models.}
   \label{fig:cil-me}
 \end{center}
\end{figure*}
\emph{\textbf{2) Parameter Efficient Expansion.}} There is a rapidly growing paradigm for parameter-efficient adaptation of pre-trained models to continually emerging classes or tasks. These approaches \cite{wang2022learning, wang2022dualprompt, smith2022coda, villa2023pivot, gao2024consistent, jung2023generating} freeze the backbone network and introduce small, learnable modules, either prompts (typically token embeddings maintained in prompt pools) or low-rank adapters to encode task-specific knowledge while minimizing catastrophic forgetting. For example, prompt-based and LoRA (Low-Rank Adaptation) \cite{hu2022lora} techniques isolate task-specific information, yielding better retention of prior knowledge than direct fine-tuning or classical continual learning methods like data replay and weight regularization. Among them, Learning to Prompt (L2P) \cite{wang2022learning} stands as the pioneering work that formulates the problem of learning new tasks as training small prompt parameters attached to a pre-trained frozen model. To further improve the plasticity of the model, CODA-Prompt \cite{smith2022coda} introduces an attention-based end-to-end key-query scheme, wherein a collection of input-conditioned prompts is learned. Wang \etal~\cite{wang2023hierarchical} proposed a hierarchical decomposition prompt framework to explicitly optimize the within-task prediction, task-identity inference, and task-adaptive prediction with task-specific prompts. Guo et al., \cite{guo2024desire} designed a novel LoRA-based, rehearsal-free continual learning method that integrates new and old knowledge via dynamic representation consolidation and decision boundary refinement, achieving state-of-the-art performance.

Recent methods have further advanced the field through subspace orthogonality constraints to reduce task interference \cite{zhang2025c, chen2024dual, liang2024inflora, fu2025iap}, dynamic module allocation and rank selection to manage capacity \cite{zhang2025c, yu2025fm, lu2024adaptive}, and sophisticated routing or selection mechanisms, for instance and task agnostic adaptation. InfLoRA~\cite{liang2024inflora} achieves interference-free adaptation by designing a subspace that prevents new task updates from impacting old tasks. DualLoRA \cite{chen2024dual} introduces orthogonal and residual LoRA adapters for each task, controlled by a dynamic memory mechanism to reduce forgetting and enhance learning of new classes. Lu \etal~\cite{lu2024adaptive} adjust LoRA rank during continual learning to balance stability and plasticity. Nevertheless, it is worth noting that prompt-based methods are largely reliant on pre-trained models. As highlighted by Kim \etal~\cite{kim2023learnability}, the information leakage, both in terms of features and class labels, can exert a substantial influence on the results of those approaches.

\subsection{Extended Settings}
\label{subsec:cil-extend}

\emph{\textbf{1) Unsupervised Continual Learning.}}
Conventionally, CL abides by a pure supervised setting, \ie, the training data $\mathcal{D}_{t}=\{\bm{x}_{i}^{t}, y_{i}^{t}\}^{n_{t}}_{i=1}$ of each continual stage is entirely labeled. However, the emerging data in real scenarios are not always labeled. Recent works~\cite{joseph2022novel,Zhao_2023_ICCV_Incremental,NEURIPS2024_5ae0f7cf} begin to study the setting of unlabeled continual learning. In each continual stage, the model needs to cluster unlabeled samples and discover new classes, while mitigating forgetting the learned old classes in previous states. From this perspective, unsupervised CL could also be referred to as the continual learning version of novel class discovery as discussed in Sec.~\ref{subsec:ncd-extend}. Concurrent works FRoST~\cite{roy2022class} and NCDwF~\cite{joseph2022novel} are the first two works. They combine the self-training algorithms~\cite{han2021autonovel} in NCD for continual category discovery and the feature replay~\cite{zhu2021prototype} spirit to resist forgetting. However, the number of continual learning stages is very limited in their settings~\cite{roy2022class,joseph2022novel,liu2022residual}. Liu \etal~\cite{liu2024large} extend this task to scenarios involving multiple stages of continual learning. They discovered that by initializing with high-quality feature representations, freezing the feature extractor and training only the classifier could yield unexpectedly exceptional performance. GM~\cite{zhang2022grow} further considers outliers in the evolving training data. In the abovementioned works, the training data of each continual stage exclusively comprises samples from new classes, so their setting could be summarized as the continual version of NCD (Sec.~\ref{subsec:ncd}). However, in real-world applications, the evolving data could simultaneously contain samples from both old and new classes. As a result, the continual version of GCD (Sec.~\ref{subsec:gcd}) is more pragmatic. IGCD~\cite{Zhao_2023_ICCV_Incremental} focuses on the replay-based setting, and proposes to store class-wise support samples from density peaks. These samples could help overcome forgetting old classes when discovering new ones. Kim \etal~\cite{Kim_2023_ICCV} utilize proxy-anchor learning schemes for feature representations. They take the spirit of divide-and-conquer and explicitly split old and new classes at each continual stage. PromptCCD~\cite{cendra2024promptccd} devises a Gaussian mixture prompt pool to achieve continual category discovery with prompt learning. Ma \etal~\cite{NEURIPS2024_5ae0f7cf} study a more challenging and realistic task, namely continual generalized category discovery (C-GCD). In C-GCD, the number of stages in continual learning and the number of unlabeled novel classes are relatively large. C-GCD follows a rehearsal-free CL paradigm, in which the model has no access to previously seen samples due to privacy and memory issues. They incorporate hardness-awareness modeling and propose to replay features of more difficult classes with a higher probability. In this way, the forgetting issue is effectively mitigated, especially for those difficult classes that are more prone to confusion with other classes in the feature space. They further devise group-wise entropy regularization to overcome the model's prediction bias towards learned classes. In addition, there are a few works that explore semi-supervised~\cite{liu2024towards}, self-supervised~\cite{liu2024branch} and federated \cite{guo2024pilora} continual learning.

\emph{\textbf{2) Multimodal Large Language Models.}}
Continual learning of multimodal large language models~\cite{liu2023visual,chen2024internvl,tong2024cambrian,bai2025qwen2,chen2025eagle,liu2025llava} (MLLM) has emerged as an important research direction to enable lifelong adaptation of large vision-language models. Recent methods focus on parameter-efficient adaptations that overcome catastrophic forgetting as new tasks or domains arrive sequentially. Prompt-based approaches learn modality-specific or fused prompts that isolate task knowledge and facilitate instance or task-aware adaptation without modifying the backbone model \cite{liu2025c, zeng2024modalprompt,guo2025hide, chen2024coin, cao2024continual}. For example, Liu \etal~\cite{liu2025c} establish a multimodal vision-language continual learning benchmark and propose C-CLIP that consists of multimodal low-rank adaptation and contrastive knowledge consolidation to effectively learn new multimodal tasks while preserving the zero-shot performance of the pretrained model.
ModalPrompt~\cite{zeng2024modalprompt} introduces prototype prompts for each task and fuses information using image-text supervision. 
HiDe-LLaVA~\cite{guo2025hide} proposes a task-specific expansion and task-general fusion framework for continual instruction tuning in MLLMs.
Recently, Zhao \etal~\cite{hongbozhao25} introduced a comprehensive benchmark (MLLM-CL) for continual learning of MLLMs, consisting of two setting named domain and ability continual learning. Then, a multimodal routing mechanism is designed to select task-specific expert in complex multimodal scenarios. LLaVA-c \cite{liu2025llava} includes spectral-aware consolidation and unsupervised inquiry regularization to enable multimodal continual learning and consistently improves standard benchmark performance, matching or surpassing multitask joint learning for the first time.
These multimodal continual learning methods demonstrate superior performance over traditional strategies, while maintaining scalability and robustness to distributional shifts in practical, real-world settings.

\begin{table*}[!t]
\setlength\tabcolsep{6pt}
\centering
\renewcommand{\arraystretch}{1}
\caption{Comparisons of average accuracies of data replay methods (\%) on CIFAR100, ImageNet-Sub, and ImageNet-Full.}
\label{tab:CL-exemplar}
\resizebox{.75\linewidth}{!}{
    \begin{tabular}{llccccccccc}
        \toprule
        \multirow{2}{*}{Algorithm}&\multirow{2}{*}{Venue}& \multicolumn{3}{c}{CIFAR-100} & \multicolumn{3}{c}{ImageNet-Sub} & \multicolumn{3}{c}{ImageNet-Full } \\
        \cmidrule(lr){3-5} \cmidrule(lr){6-8} \cmidrule(lr){9-11}
        &&$T=5$ &$T=10$ &$T=25$ &$T=5$ &$T=10$ &$T=25$ &$T=5$ &$T=10$ &$T=25$\\
        \midrule
        DMC\cite{zhang2020class} & WACV 2020 & 38.20 &23.80 &– &43.07 &30.30 &– &– &– &– \\
        GD\cite{lee2019overcoming} & ICCV 2019 &56.39 &51.30 &– &58.70 &57.70 &– &– &– &– \\
        iCaRL\cite{Rebuffi2017iCaRLIC} & CVPR 2017 &57.12 &52.66 &48.22 &65.44 &59.88 &52.97 &51.50 &46.89 &43.14 \\
        BiC\cite{Wu2019LargeSI} & CVPR 2019 &59.36 &54.20 &50.00 &70.07 &64.96 &57.73 &62.65 &58.72 &53.47 \\
        TPCIL\cite{tao2020topology} & ECCV 2020 &65.34 &63.58 &– &76.27 &74.81 &– &64.89 &62.88 &– \\
        UCIR\cite{Hou2019LearningAU} & CVPR 2019 &63.17 &60.14 &57.54 &70.84 &68.32 &61.44 &64.45 &61.57 &56.56 \\
        WA\cite{ZhaoXGZX20} & CVPR 2020 &61.70 &56.37 &50.78 &71.26 &64.99 &53.61 &56.69 &52.35 &44.58\\
        PODnet\cite{Douillard2020SmallTaskIL} & ECCV 2020 & 64.83 &63.19 &60.72 &75.54 &74.33 &68.31 &66.95 &64.13 &59.17 \\
        +DDE\cite{hu2021distilling} & CVPR 2021 &65.42 &64.12 &– &76.71 &75.41 &– &66.42 &64.71 &–\\
        +AANets\cite{liu2021adaptive} & CVPR 2021 &66.31 &64.31 &62.31 &76.96 &75.58 &71.78 &67.73 &64.85 &\underline{61.78} \\
        +MRDC\cite{wang2021memory} & ICLR 2022 &– &– &– &\textbf{78.08} &\underline{76.02} &\underline{72.72} &\textbf{68.91} &\underline{66.31} &– \\
        +CwD\cite{shi2021mimicking} & CVPR 2022 & 67.44 &64.64 &62.24 &76.91 &74.34 &67.42 & 58.18 &56.01 &–\\
        Mnemonics\cite{liu2020mnemonics} & CVPR 2020 &63.34 &62.28 &60.96 &72.58 &71.37 &69.74 &64.54 &63.01 &61.00 \\
        +AANets\cite{liu2021adaptive} & CVPR 2021 &67.59 &65.66 &63.35 &72.91 &71.93 &70.70 &65.23 &63.60 &61.53\\
        RMM\cite{liu2021rmm} & NeurIPS 2021 & \underline{68.42} &\underline{67.17} &\textbf{64.56} &73.58 &72.83 &72.30 &65.81 &64.10 &\textbf{62.23}\\
        DER\cite{yan2021dynamically} & CVPR 2021 & \textbf{72.60} & \textbf{72.45} &– &– &\textbf{77.73} &– &– &– &–\\
        SSIL\cite{ahn2021ss} & ICCV 2021 &63.02 &61.52 &58.02 &– &– &– &– &– &– \\
        AFC\cite{Kang2022CVPR} & CVPR 2022 &66.49 &64.98 &\underline{64.06} &76.87 &75.75 &\textbf{73.34} &\underline{68.90} &\textbf{67.02} &– \\
        DMIL\cite{Tang2022CVPR} & CVPR 2022 &68.01 &66.47 &– &\underline{77.20} &76.76 &– &– &– &– \\
        \bottomrule
    \end{tabular}}
\end{table*}
\subsection{Theoretical Study of Continual Learning}
\label{subsec:cl-theory}
With substantial advancements in the empirical methods, recent works also explore the theoretical aspect of continual learning. One line of work considers the setting where all tasks share the same global minimizer. Evron \etal~\cite{evron2022catastrophic} investigated fitting an overparameterized linear regression model for a sequence of tasks with varying input distributions, proving an upper bound of $T^2 \min\left\{1/\sqrt{k}, d/k\right\}$ on the forgetting, where $T$ tasks in $d$ dimensions are presented cyclically for $k$ iterations. Further, Evron \etal~\cite{evron2023continual} studied continual learning of separable linear classification tasks with binary labels, deriving upper bounds on the forgetting effect. Ding \etal~\cite{dingunderstanding} focused on the CL of linear regression model via SGD and demonstrated that training tasks with larger eigenvalues in their data covariance matrices later in the sequence increase forgetting when data size is sufficiently large, while an appropriate step size mitigates forgetting. Peng \etal~\cite{peng2023ideal} introduced the concept of an ideal continual learner and established its correlation with existing regularization based, data replay based and expansion based methods. The role of Mixture-of-Experts in CL for linear regression tasks under overparameterized regimes has been investigated \cite{li2024theory}.

Another line of work considers a more general setting where there is no shared global minimize among the continually learned tasks. Within the Neural Tangent Kernel framework \cite{bennani2020generalisation, doan2021theoretical}, the generalization error and forgetting phenomena for orthogonal gradient descent methods \cite{farajtabar2020orthogonal} have been analyzed. 
The influence of task similarity on forgetting has been respectively explored for knowledge distillation based methods \cite{lee2021continual, asanuma2021statistical} under overparameterized regimes. Lin \etal~\cite{lin2023theory} provided the explicit form of the forgetting and generalization error for CL of linear regressions with an arbitrary number of tasks, and demonstrated how overparameterization, task
similarity, and task ordering affect both forgetting
and generalization error. For example, the authors showed that generalization error consistently decreases as tasks become more similar while forgetting can even diminish when tasks are less similar. Deng \etal~\cite{deng2025} theoretically demonstrated that sequential rehearsal outperforms standard concurrent rehearsal in mitigating catastrophic forgetting and improving generalization when tasks are less similar in continual learning. 
Additionally, Kim \etal~\cite{kim2022theoretical} established a link between task and class continual learning, further providing proof of the learnability of class continual learning \cite{kim2023learnability}.

Despite significant advancements, existing theoretical investigations suffer from several fundamental limitations. Firstly, nearly all prior works \cite{evron2022catastrophic,evron2023continual,dingunderstanding,peng2023ideal,lin2023theory} rely on a naive locally i.i.d. assumption, \ie, the data in each task is required to be independent and identically distributed (i.i.d.) to simplify the analysis, which may be difficult to verify or guarantee in non-stationary nature of real-world data streams, specifically when continually learning and interacting with the dynamic, open environments \cite{zhang2020towards}. Secondly, these studies often impose stringent conditions such as persistent excitation to ensure feature diversity or assume standard Gaussian noise, both of which are difficult to validate in practice. Thirdly, most research focuses on linear regression models, ignoring nonlinear cases (\eg, ReLU/sigmoid activations). Besides, the overparameterized regime assumed in some works lacks applicability to large-scale datasets. A recent work \cite{zhu2025global} presented a comprehensive theoretical analysis of continual learning for general convex models, and established almost sure convergence results under general data conditions. Moreover, the authors provided the convergence rates for forgetting and regret without excitation assumptions, and derived non-asymptotic error bounds for parameter estimation and CL performance for finite datasets. In future research, a number of theoretical problems remain to be explored, \eg, the convergence properties or estimation error bounds of continual learning with no shared global minimizer, and theoretical investigation for continual learning with DNNs.

\begin{table}[!t]
\setlength\tabcolsep{3pt}
\centering
\renewcommand{\arraystretch}{1.1}
\caption{Comparisons of the average accuracies (\%) of non-exemplar based continual learning methods.}
\label{tab:CL-nonexemplar}
\resizebox{\linewidth}{!}{
\begin{tabular}{llccccccc}
\toprule
\multirow{2}{*}{Algorithm} & \multirow{2}{*}{Venue} & \multicolumn{3}{c}{CIFAR-100} & \multicolumn{3}{c}{Tiny-ImageNet} & \multicolumn{1}{c}{ImageNet-Sub} \\
\cmidrule(lr){3-5} \cmidrule(lr){6-8} \cmidrule(lr){9-9}
 & & $T=5$ & $T=10$ & $T=20$ & $T=5$ & $T=10$ & $T=20$ & $T=10$ \\
\midrule
LwF-MC \cite{Li2018LearningWF}  & ECCV 2016 & 33.4 & 26.0 & 19.7 & 34.9 & 21.4 & 13.7 & 35.8 \\
LwM \cite{DharSPWC19}  & CVPR 2019 & 39.6 & 30.2 & 20.5 & 37.3 & 20.5 & 12.6 & 32.6 \\
MUC \cite{liu2020more}  & ECCV 2020 & 49.3 & 36.0 & 29.0 & 37.5 & 26.3 & 21.6 & -- \\
CCIL \cite{Li2018LearningWF} & ICME 2021 & 60.8 & 43.6 & 38.1 & 36.7 & 27.6 & 16.3 & 41.1 \\
UCIR-DF \cite{Hou2019LearningAU} & CVPR 2019 & 57.8 & 48.7 & -- & -- & -- & -- & -- \\
PODNet-DF \cite{Douillard2020SmallTaskIL} & ECCV 2020 & 56.9 & 52.6 & -- & -- & -- & -- & -- \\
ABD \cite{smith2021always} & ICCV 2021 & 62.4 & 59.0 & -- & 44.6 & 41.6 & -- & -- \\
R-DFCIL\cite{gao2022r} & ECCV 2022 & 64.8 & 61.7 & -- & 48.9 & 47.6 & -- & -- \\
IL2A \cite{zhu2021class} & NeurIPS 2021 & 66.2 & 58.2 & 58.0 & 47.2 & 44.7 & 40.0 & 58.0 \\
PASS \cite{zhu2021prototype} & CVPR 2021 & 63.8 & 59.9 & 58.1 & 49.5 & 47.2 & 42.0 & 62.1 \\
PASS++ \cite{zhu2024pass++} & TPAMI 2025 & \underline{69.1} &66.5 &\underline{64.3} &54.1 &53.1 &49.7 &71.9 \\
SSRE \cite{zhu2022self} & CVPR 2022 & 65.9 & {65.0} & {61.7} & 50.4 & 48.9 & 48.2 & 67.7 \\
SDC \cite{yu2020semantic} & CVPR 2020 & 66.2 & 62.7 & 59.2 & 53.3 & 50.5 & 48.8 & 68.6 \\
Fusion \cite{toldo2022bring} & CVPR 2022 & {66.9} & 64.8 & 61.5 & {54.2} & {52.6} & \underline{50.2} & {69.3} \\
FeTrIL \cite{petit2023fetril} & WACV 2023 & 67.6 & \underline{66.6} & 63.5 & \underline{55.4} & \underline{54.3} & 53.0 & {71.9} \\
Eucl-NCM & NeurIPS 2023 & 64.8 & 64.6 & 61.5 & 54.1 & 53.8 & \underline{53.6} & \underline{72.0} \\
FeCAM \cite{goswami2023fecam} & NeurIPS 2023 & \textbf{70.9} & \textbf{70.8} & \textbf{69.4} & \textbf{59.6} & \textbf{59.4} & \textbf{59.3} & \textbf{78.2} \\
\bottomrule
\end{tabular}
}
\end{table}

\subsection{Evaluation of Continual Learning}
\label{subsec:cl-evaluation}
In continual learning, performance is typically assessed by last accuracy and average accuracy. 
Given a dataset containing \(M\) classes, one can divide all classes into \(T\) disjoint tasks for the model to learn sequentially. Each task contains the same number of classes (\(M/T\), assuming \(M\) is divisible by \(T\)). However, motivated by practical scenarios where the initial task often contains more classes, many works assign half of the classes to the initial task and evenly divide the remaining half into \(T\) subsequent tasks. Each of these tasks, therefore, contains \(M/(2T)\) classes, resulting in a total of \(T+1\) tasks. For instance, when applied to CIFAR-100, the initial task would include 50 classes, and the remaining 50 classes would be split into 10 tasks of 5 classes each. Then, at the $t$-th continual stage, the last accuracy $a_t$ is defined as the top-1 accuracy of all learned classes at the final continual stage, and the average accuracy is computed as $A_{t}=\frac{1}{t}\sum_{i=1}^{t}a_{i}$. Following the above evaluation protocols~\cite{Douillard2020SmallTaskIL, gao2022r}, Table \ref{tab:CL-exemplar} and Table \ref{tab:CL-nonexemplar} provide comprehensive results on common benchmark datasets such as CIFAR-100~\cite{krizhevsky2009learning}, Tiny-ImageNet~\cite{le2015tiny}, and ImageNet \cite{deng2009imagenet} under different continual stages. As can be seen, data replay methods (Table \ref{tab:CL-exemplar}) typically achieve good performance, where imbalance calibration is important. However, recent developed non-exemplar methods (Table \ref{tab:CL-nonexemplar}) like PASS++ \cite{zhu2024pass++} and FeCAM \cite{goswami2023fecam} can also yield strong continual learning performance.

\section{Future Directions}
\label{sec:future}

\begin{figure*}[t]
	\begin{center}
		\centerline{\includegraphics[width=\textwidth]{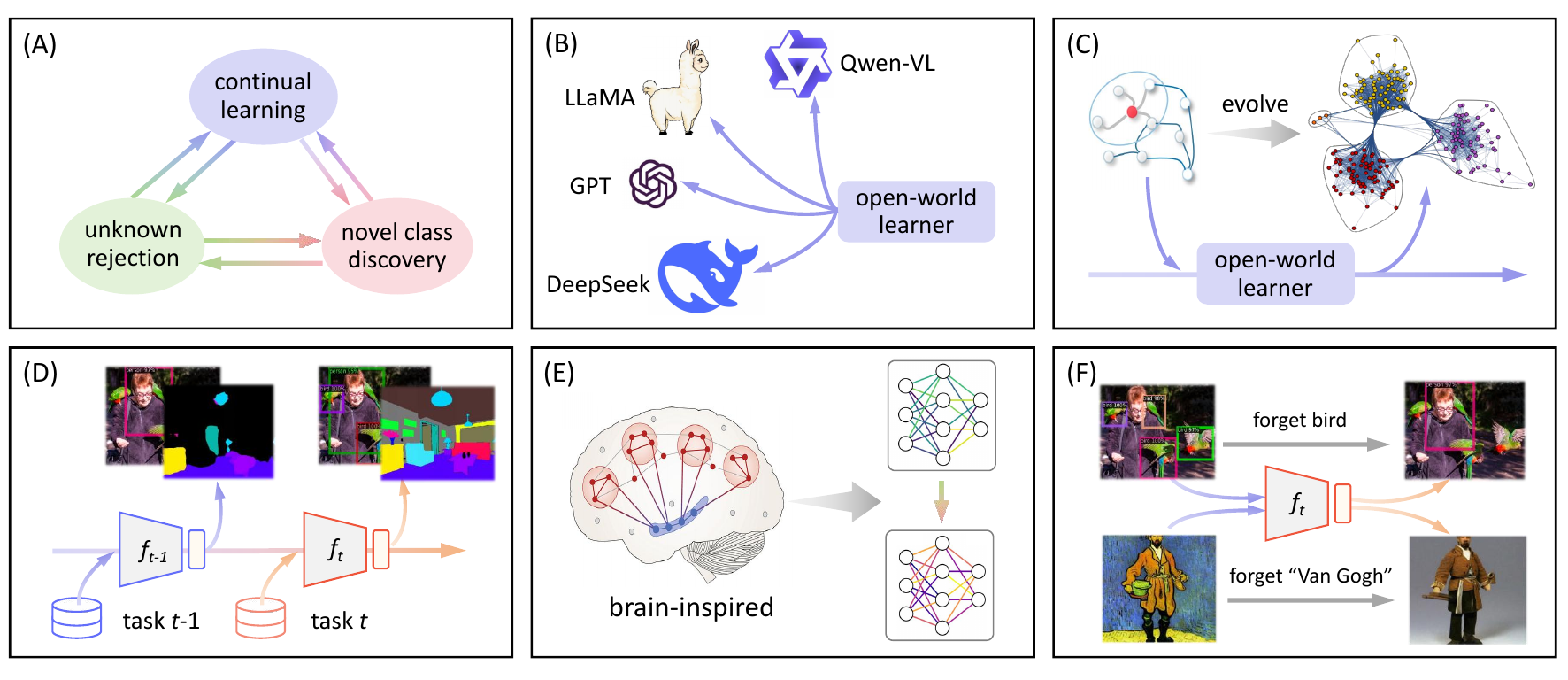}}
		\vskip -0.1in
		\caption{Illustrations of future research directions. (A) Unified open-world learner. (B) OWL for large models. (C) OWL for structured data with graph neural networks. (D) OWL beyond classification, such as object detection and segmentation. (E) Brain-inspired OWL models that mimic biological learning systems. (F) Open-world unlearning actively and selectively forgetting or erasing specific knowledge.}
		\label{fig:future}
	\end{center}
	\vskip 0.05in
\end{figure*}

Open-world learning (OWL) is an active and long-term research topic, it holds transformative potential for developing adaptive artificial intelligence systems that can operate in unpredictable scenarios, from autonomous robotics to personalized healthcare. Over the past decade, substantial strides have been made in endowing systems with the ability to manage the uncertainty inherent in open environments, discover novel concept, and continually update their knowledge. Yet, as OWL moves toward broader real-world deployment, there are a number of critical open directions worthy of further investigation. In this section, we briefly outline a few promising research directions that enable OWL in a unified framework and more complex scenarios (illustrated in Fig.~\ref{fig:future}), \emph{\eg} structured data and applications such as detection, segmentation, \etc. Besides, additional directions considering brain-inspired OWL and machine unlearning are also discussed.

\emph{\textbf{1) Unified Open-world Learning.}}
A robust open-world learner must continuously adapt through ongoing interaction with its environment, seamlessly integrating classification, rejection, discovery, and continual learning within a single pipeline. During deployment, the model should accurately recognize classes it has encountered so far, while simultaneously rejecting both misclassified examples from known categories and out-of-distribution samples representing unseen classes. Instead of treating rejection and adaptation as isolated tasks, the learner should maintain a buffer of rejected instances and analyze them to identify genuine novelties. By assigning provisional labels to these new classes, the model effectively expands its set of known categories and prepares them for subsequent training. Crucially, as the learner incorporates these discoveries into its knowledge base, it must preserve discrimination performance on previously learned classes and avoid catastrophic forgetting. Existing approaches often address one aspect of this process in isolation. For example, continual learning benchmarks typically assume that new training data are clean and manually labeled according to discrete class sets, whereas real-world applications require autonomous discovery and labeling from an unlabeled stream of mixed samples. Moreover, the dynamic interplay among rejection, novelty discovery, and continual adaptation remains poorly understood. Although recent work~\cite{kim2023learnability} has shown that OOD detection techniques can enhance continual learning by filtering novel instances, the shifting decision boundaries that result from accumulating knowledge may in turn affect rejection behavior over time. To move toward truly autonomous, scalable, and reliable open-world learners, it is essential to study the interactions among rejection, discovery, and continual adaptation in concert and to develop unified frameworks that can address all three challenges simultaneously.

\emph{\textbf{2) Open-world Learning of Large Models.}}
Foundation models such as LLaMA \cite{touvron2023llama}, DeepSeek \cite{liu2024deepseek}, LLaVA \cite{liu2024improved}, Qwen \cite{yang2024qwen2} and the GPT4 \cite{OpenAI2023GPT4TR} have greatly advanced zero‑shot and few‑shot capabilities in both language and multimodal understanding. However, these models exhibit limitations when deployed in truly open‑world environments where inputs may diverge substantially from their training distributions. One significant challenge is hallucination, which refers to the output of plausible but incorrect or fabricated information \cite{zhang2023siren}. Within an open‑world learning framework, the detection of hallucinations corresponds directly to the problem of unknown rejection, requiring models to withhold responses when confidence is low or when input patterns deviate from established distributions \cite{ren2022out}.
Another challenge derives from the evolving nature of real‑world knowledge. Without mechanisms for ongoing adaptation, even the most advanced large models can rapidly provide outdated or inaccurate information. To this end, continual pretraining and continual instruction tuning is needed. Specifically, Continual pretraining involves periodically updating model parameters using newly available corpora, thereby ensuring that the model reflects the latest advancements in fields such as science, law and software development \cite{jang2021towards}. Continual instruction tuning refines a model’s capacity to follow updated user directives and adhere to current protocols \cite{guo2025hide, zeng2024modalprompt}. For example, a large model that undergoes continual instruction tuning on recent multimedia search queries will interpret novel query formulations more accurately. To develop OWL methods for large models, we must consider the difference between traditional discriminative models and large generative models, which have more complex generation objectives and diverse forms of task. Furthermore, incorporating user feedback loops, where corrected or clarified responses are incorporated into subsequent instruction‑tuning cycles, can enable models to self‑improve over time.
With open-world learning, the next generation of large models will be better equipped to deliver robust, adaptive performance in complex, dynamic scenarios.

\emph{\textbf{3) Open-world Learning of Structured Data.}}
Most open-world learning methods assume that inputs are sampled independently. In many real-world scenarios, however, data exhibit rich structure and interdependence, as seen in traffic networks, social graphs, and molecular interaction maps. Existing open-world approaches struggle with such settings because they focus on isolated instances and overlook the topological relationships that bind samples together. There have been some studies that attempt to improve the OWL ability for structured data learning with graph neural networks. Stadler \etal~\cite{stadler2021graph} proposed a Bayesian graph neural network architecture capable of detecting out-of-distribution nodes, while Wu \etal~\cite{wuenergy} developed an energy-based discriminator to reject unknown inputs in graph-structured data. Liu \etal~\cite{liu2021overcoming} introduced a topology-aware weight-preservation strategy that explicitly models graph aggregation and propagation, reducing catastrophic forgetting of previously learned relationships during continual learning. Zhang \etal~\cite{zhang2022cglb} further contributed a comprehensive benchmark for continual graph learning, including both node-level and entire-graph tasks. Open-world learning on structured data is an emerging research area, and future work should leverage a deeper understanding of structured data’s topological properties to develop topology-aware open-world learning methods that can robustly discover and adapt to new knowledge in complex, interconnected scenarios.

\emph{\textbf{4) Open-world Learning Beyond Classification.}}
Existing works mainly focused on image classification tasks. Recently, there has been noticeable attention in addressing open-world learning challenges in other applications, such as object detection \cite{wang2023detecting, joseph2021incremental} and semantic segmentation \cite{zhang2022mining, xiao2023endpoints, xu2024dual}. In image classification, a single image typically only contains one object. In contrast, object detection and semantic segmentation tasks typically involve input images that contain multiple objects, and some of them are labeled during the initial learning stage, while others should be classified as unknown. This implies that many unknown classes would have already been seen by the object detector.
During the continual learning process, information (labels) about these new classes within the background becomes available, the model should incrementally learn them to incorporate novel knowledge \cite{joseph2021towards}. Therefore, the advances in open-world classification cannot be trivially adapted to open-world object detection or segmentation, and it is important to leverage background information and explore feature representations that could facilitate the future learning of new classes. Besides, OWL is also important for the natural language processing domain, where out-of-vocabulary words or novel concept vocabulary may emerge over time. To design effective algorithms for those tasks beyond classification, it is helpful to combine existing methods with task-specific domain knowledge.

\emph{\textbf{5) Brain-inspired Open-world Learning.}}
Compared to DNN-based artificial intelligence systems, biological organisms like humans naturally have the strong ability of open-world learning, enabling them to interact with the environment throughout their lifetime \cite{kudithipudi2022biological}. For example, humans can be aware of when they are likely to be wrong and then refuse to make a high-risk decision; they can discover novel knowledge based on what they have learned previously, and last but not least, humans excel at learning from a dynamically changing environment incrementally without suffering from catastrophic forgetting In light of this, some works \cite{van2020brain, zeng2019continual, wang2023incorporating, shen2018incremental, wang2021triple} aim to design brain-inspired open-world machine learning models. For example, van de Ven \etal~\cite{van2020brain} proposed a brain-inspired feature replay method that uses the original classification model itself to generate features of old classes, avoiding using additional generative models and showing promising performance for CL. Recently, inspired by biological learning systems, Wang \etal~\cite{wang2023incorporating} built a continual model with the ability of stability protection and active forgetting. Inspired by human memory, Liu \etal~\cite{liu2025semi} introduced a biomimetic continual learning framework that combines semi-parametric memory with a wake–sleep consolidation mechanism, enabling deep neural networks to acquire new tasks without forgetting prior knowledge.
Building on these advances, future research in brain-inspired open-world learning should focus on developing mechanisms for adaptive confidence estimation to gauge when models should seek feedback \cite{pouget2016confidence}, fostering active exploration and curiosity-driven modules to autonomously discover novel information \cite{modirshanechi2023curiosity}, integrating multimodal data streams to form richer knowledge representations. Besides, drawing inspiration from the brain’s sparse, it is important to create energy-efficient edge implementations and adaptation in ever-changing environments, ultimately narrowing the gap between artificial and natural intelligence.

\emph{\textbf{6) Open-world Unlearning.}}
Most of the focus in machine learning is on training models to acquire new knowledge, \eg, continual learning aims to incorporate new knowledge without forgetting previously learned knowledge. On the contrary, there are scenarios where machine unlearning may be required for various reasons: (1) Sensitive information removal. In some cases, machine learning models may inadvertently memorize sensitive or private information from the training data \cite{dilmaghani2019privacy, gandikota2023erasing}. Unlearning can be used to erase such information from the model's memory to ensure data privacy and compliance with regulations. (2) Bias mitigation. If a model exhibits undesirable biases in its predictions \cite{wang2020towards}, unlearning can be employed to eliminate these biases or align the model better with desired behaviors by removing some patterns learned during training. (3) Selective forgetting. Unlearning allows models to discard some outdated information to leave more room for learning new tasks. This is different from the well-known catastrophic forgetting \cite{Goodfellow2014AnEI, mccloskey1989catastrophic}, which may forget important information that we want to memorize. 
Indeed, machine unlearning is challenging because it involves determining which parts of knowledge to remove, developing algorithms for knowledge erasure, and ensuring that unlearning does not adversely affect the model's overall performance \cite{xu2023machine, zhang2023forgetmenot}.
Therefore, it is a long-term research goal to perform open-world learning and unlearning flexibly and simultaneously. To reach this goal, future research must address several open problems. First, we need principled frameworks to quantify and localize the knowledge to be unlearned without requiring full retraining, especially under resource constraints. In light of this, Zhao \etal~\cite{zhao2025practical, zhao2024continual} defined the problem of continual forgetting, which focuses on selective and progressive removal of information from pre-trained vision models, and propose a method to effectively forget targeted knowledge with minimal impact on retained knowledge.
Second, it is crucial to design unlearning methods that are robust to the uncertainty and dynamics of open-world environments, where new classes may appear and old ones may evolve or disappear. Third, the interplay between learning and unlearning processes should be better understood, \eg, how they influence each other and how to balance them effectively to maintain model stability and adaptability.

\section{Conclusion}
\label{sec:conclusion}
Open-world learning stands as a critical frontier in the field of artificial intelligence, particularly as AI systems increasingly operate in complex, real-life environments, and engage with humans and automated systems. The essence of open-world learning lies in the ability to detect and adapt to new, diverse and ever-changing scenarios, like a self-motivated learner. 
The paper presents a comprehensive taxonomy, reviews numerous papers, and provides
insights into open-world learning, with an emphasis on techniques concerning unknown rejection, novel class discovery, and continual learning. It also offers performance comparison. 
Finally, we outlined several challenging yet important research directions that merit substantial research endeavors in the future.

\bibliographystyle{unsrt}
\bibliography{refer}

\end{document}